\documentclass{article}

\PassOptionsToPackage{numbers, compress}{natbib}

\usepackage[preprint]{neurips_2026}

\usepackage[utf8]{inputenc} 
\usepackage[T1]{fontenc}    
\usepackage{hyperref}       
\usepackage{url}            
\usepackage{booktabs}       
\usepackage{amsfonts}       
\usepackage{nicefrac}       
\usepackage{microtype}      
\usepackage{xcolor}         

\usepackage{amsmath}
\usepackage{amssymb}
\usepackage{graphicx}
\usepackage{float}
\usepackage{enumitem}
\usepackage{tabularx}
\usepackage{mathtools}

\definecolor{blueVar}{RGB}{1, 115, 178}
\definecolor{greenVar}{RGB}{2, 158, 115}
\definecolor{orangeVar}{RGB}{222, 143, 5}
\definecolor{purpleVar}{RGB}{204, 120, 188}
\definecolor{redVar}{RGB}{213, 94, 0}
\definecolor{lightblueVar}{RGB}{86, 180, 233}
\definecolor{greyVar}{RGB}{128, 128, 128}

\newcommand{\synclr}[1]{\textcolor{lightblueVar}{#1}} 
\newcommand{\mlpclr}[1]{\textcolor{orangeVar}{#1}}    
\newcommand{\memclr}[1]{\textcolor{blueVar}{#1}}      
\newcommand{\outclr}[1]{\textcolor{purpleVar}{#1}}    

\title{Scaling Laws and Tradeoffs\break in Recurrent Networks of Expressive Neurons}

\author{
  Aaron Spieler\textsuperscript{1,2} \quad
  Georg Martius\textsuperscript{1,3} \quad
  Anna Levina\textsuperscript{1,2}\\[0.5em]
  \textsuperscript{1}University of Tübingen, Germany\\
  \textsuperscript{2}Max Planck Institute for Biological Cybernetics, Tübingen, Germany\\
  \textsuperscript{3}Max Planck Institute for Intelligent Systems, Tübingen, Germany
  \vspace{-1em}
}

\begin{document}

\maketitle

\begin{abstract}

Cortical neurons are complex, multi-timescale processors wired into recurrent circuits, shaped by long evolutionary pressure under stringent biological constraints. Mainstream machine learning, by contrast, predominantly builds models from extremely simple units, a default inherited from early neural-network theory.
We treat this as a normative architectural question. How should one split a fixed parameter budget $P$ between the number of units $N$, per-unit effective complexity $k_e$, and per-unit connectivity $k_c$? What controls the optimal allocation? This calls for a model in which per-unit complexity can be tuned independently of width and connectivity. 
Accordingly, we introduce the ELM Network, whose recurrent layer is built from Expressive Leaky Memory (ELM) neurons, chosen to mirror functional components of cortical neurons: multi-timescale memory, structured synaptic integration, and nonlinear internal computation. The architecture allows for individually adjusting $N$, $k_e$, and $k_c$ and trains stably across orders of magnitude in scale.
We evaluate the model on two qualitatively different sequence benchmarks: the neuromorphic SHD-Adding task and Enwik8 character-level language modeling. 
Performance improves monotonically along each of the three axes individually. Under a fixed budget, a clear non-trivial optimum emerges in their tradeoff, and larger budgets favor both more numerous \emph{and} more complex neurons. A closed-form information-theoretic proxy captures these tradeoffs and attributes the diminishing returns at two ends to: per-neuron signal-to-noise saturation and across-neuron redundancy.
Connectivity enters as a related mechanism that helps neurons learn distinct signals.
A hyperparameter sweep spanning three orders of magnitude in trainable parameters traces a near-Pareto-frontier scaling law consistent with the proxy, mapping the budget-constrained tradeoff surface between unit count, unit complexity, and connectivity.
This suggests that the simple-unit default in ML is not obviously optimal once this tradeoff surface is probed, and offers a normative lens on cortex's reliance on complex spatio-temporal integrators.

\end{abstract}

\section{Introduction}

\begin{figure*}[ht]
    \centering
    \includegraphics[width=1.0\linewidth]{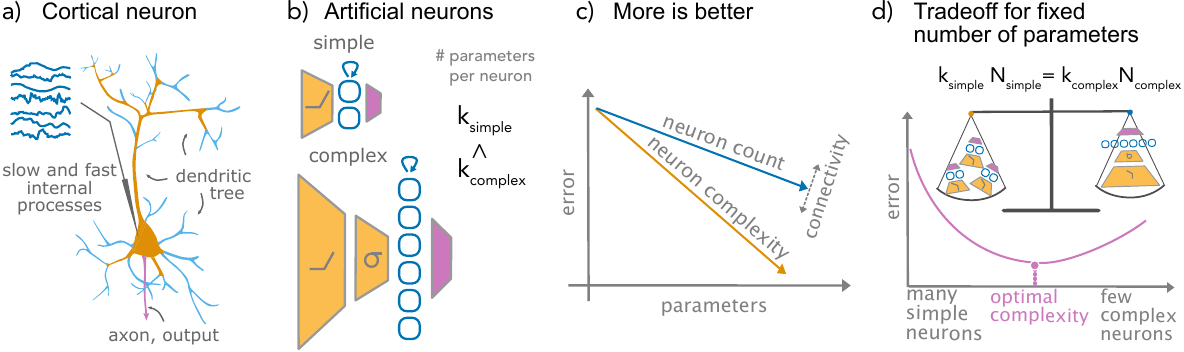}
    \vspace{-0.75em}
    \caption{\textbf{Complexity of cortical neurons motivates the search for the optimal complexity of a unit in parameter-constrained recurrent networks.}
    \textbf{a)} Cortical neurons combine complex dendritic structure with rich internal dynamics, making them powerful spatio-temporal processing units.
    \textbf{b)} Using ELM neurons \citep{spieler2024the}, from simple integrators with two memory units and a few parameters $k_{\mathrm{simple}}$ to large models exceeding the complexity of cortical neurons with many parameters $k_{\mathrm{complex}} \gg k_{\mathrm{simple}}$. 
    \textbf{c)} Recurrent networks with more neurons or neurons of larger complexity are expected to perform better; connectivity affects the slope.
    \textbf{d)} Under a fixed parameter budget, there is a non-trivial tradeoff between network size and neuron complexity, which our theory explains.}
    \label{fig:graphical_motivation}
    \vspace{-1.0em}
\end{figure*}

Cortical neurons are sophisticated spatio-temporal processors. Their dendritic morphology, active conductances, and multi-timescale internal dynamics support nonlinear integration well beyond a point-wise nonlinearity \citep{larkum2022dendrites,aizenbud2026dendritic,poirazi2020illuminating} (Fig.~\ref{fig:graphical_motivation}a). Mainstream machine learning, in contrast, builds models from simple units, a default inherited from early neural-network theory \citep{Chakraverty2019} and largely unchanged across modern recurrent, state-space, and attention-based architectures \citep{vaswani2017attention,gu2023mamba,beck2024xlstm}.

The two fields have increased model complexity along separate axes. Computational neuroscience has refined the single-neuron model to match the experimental observations along two strands that are rarely combined. One strand enriches the dynamical machinery at the soma, layering subthreshold currents, adaptation, spike-frequency dependence, and energy or metabolic constraints onto an otherwise point-like unit \citep{izhikevich2004model,brette2005adaptive,gerstner2014neuronal,bellec2018long,fardet2020simple}. The other formalizes spatial integration carried out in the dendritic tree, from the canonical two-layer model of pyramidal computation to active dendritic compartments and structured input integration \citep{larkum2022dendrites,poirazi2003pyramidal,jadi2014augmented,gidon2020dendritic,ujfalussy2018global,stuart2015dendritic}. Only a few trainable neuron models have attempted to flexibly capture both directions, with most efforts focusing on the single-neuron setting \citep{beniaguev2021single, spieler2024the}.\looseness-1

Machine learning has mostly moved along a different axis: instead of enriching computation within units, modern architectures keep units simple and scale by structuring interactions---through convolution, gating, attention, and connectivity \citep{lecun1998gradient,hochreiter1997long,vaswani2017attention,gu2022efficiently,gu2023mamba}---while stacking layers wide and deep. To assess how much additional per-unit complexity is beneficial for efficient computation and to relate these findings to the complexity of cortical neurons \citep{poirazi2003pyramidal,gidon2020dendritic,beniaguev2021single,aizenbud2026dendritic}, we require a setup in which unit complexity is a controllable parameter within an end-to-end trained network. Scaling laws provide a natural framework for this question: they describe how performance improves as resources increase along specific axes and can inform how to allocate a fixed budget across design choices \citep{kaplan2020scaling,hoffmann2022training}. Prior work has primarily studied scaling with model size, data, and compute, often observing power-law trends with parameter-specific exponents. By contrast, scaling with per-unit complexity remains largely unexplored, despite biological neurons suggesting it may be a relevant axis.

To address this gap, we introduce the ELM Network, a wide recurrent model whose per-unit complexity can be adjusted along neuron-relevant dimensions and that trains stably across three orders of magnitude in trainable parameters, building on the Expressive Leaky Memory neuron \citep{spieler2024the}. To avoid scaling artifacts from hardware accelerators, we use trainable parameters under a fixed training setup as a cross-disciplinary budget. We sweep the three knobs: number of neurons $N$, internal neuron complexity $k_\text{e}$, and connectivity parameters $k_\text{c}$ on two contrasting sequence benchmarks: the neuromorphic SHD-Adding task \citep{cramer2020heidelberg, spieler2024the} and Enwik8 character-level language modeling. Performance grows monotonically with each axis taken alone, but under a fixed budget, a clear, nontrivial optimum emerges in their joint trade-off (Fig.~\ref{fig:graphical_motivation}c,d), and the location of that optimum shifts toward both larger and more complex units as the budget increases. We complement these empirical results with a closed-form information-theoretic model whose interpretable parameters track distinct architectural axes in our experiments and qualitatively reproduce the observed trade-offs under a joint fit. These findings undercut the assumption that simple units are an obviously optimal default in machine learning and lend a normative reading to cortex's reliance on complex, multi-timescale units.

Our contributions are the following:
\vspace{-0.5em}
\begin{enumerate}\setlength{\itemsep}{1pt}\setlength{\parskip}{1pt}\setlength{\parsep}{1pt}
    \item The ELM Network as a robust and scalable model system for recurrent computation with expressive neurons, in which neuron count, complexity, and connections vary independently.
    \item Experimental scaling and tradeoff results for such networks showing monotonic gains along each axis alone, but non-trivial optima under fixed parameter budget.
    \item A compact information-theoretic framework for explaining the tradeoffs in such networks in terms of single-neuron signal-to-noise and population activity redundancy.
\end{enumerate}

\clearpage
\section{The Model Architecture}
\label{sec:model_architecture_section}

\begin{figure*}[ht]
    \centering
    \includegraphics[width=1.0\linewidth]{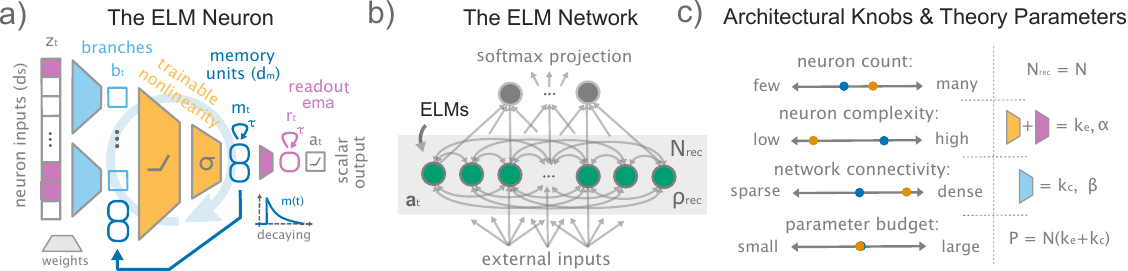}
    \vspace{-0.75em}
    \caption{\textbf{A stable and flexible model system for studying scaling and tradeoffs in recurrent networks of expressive neurons.}
    \textbf{a)} The modified Expressive Leaky Memory (ELM) neuron.
    \textbf{b)} ELM neurons are assembled as ELM Network, a doubly recurrent sequence model whose computational core is a wide recurrent hidden layer in which each neuron itself is recurrent; followed by a smaller readout layer and output projection.
    \textbf{c)} (left) The main study axes as explicit architectural knobs (right) Corresponding theory parameters: neuron count to $N$, effective neuron complexity to $(k_e,\alpha)$, network connectivity structure to $(k_c,\beta)$, and total budget to $P = N(k_e+k_c)$. We count $k_e = \#w_p + \#w_r$ and $k_c = \#w_s$. Theory $\alpha$, $\beta$ are introduced in Section~\ref{sec:information_theory_framework}. Model parameters in Appendix. Tab. \ref{tab:param_table}.}
    \label{fig:model_figure}
\end{figure*}

We assemble Expressive Leaky Memory (ELM) neurons~\citep{spieler2024the} into a recurrent architecture for sequential processing, comprising three nested hierarchies: neuron, layer, and network (Fig.~\ref{fig:model_figure}).

\vspace{-0.3em}
\paragraph{The ELM Neuron.}
Each ELM neuron is a recurrent cell with $d_m$ leaky memory units and a single scalar output; schematic overview in Figure~\ref{fig:model_figure}a. At each time step, the neuron \textbf{(i)}~weights its synaptic inputs  $\boldsymbol{z_t} \in \mathbb{R}^{d_s}$ using $\synclr{\boldsymbol{w_s}} \in \mathbb{R}^{d_s}$ ($d_s = d_\text{tree} \cdot d_\text{branch}$), groups them into $d_\text{tree}$ \synclr{branches}, and sums within each to get $\synclr{\boldsymbol{b_t}} \in \mathbb{R}^{d_\text{tree}}$; \textbf{(ii)}~produces a bounded memory \mlpclr{update} proposal $\mlpclr{\Delta\mathbf{m}_t} \in \mathbb{R}^{d_m}$ by passing the branch activations and the decayed memory state through a nonlinear MLP with $l_\text{mlp}$ hidden layers of width $d_\text{mlp}$; \textbf{(iii)}~updates the \memclr{memory} units $\memclr{\boldsymbol{m}_t} \in \mathbb{R}^{d_m}$ via leaky integration with per-unit timescales $\memclr{\boldsymbol{\tau}_m} \in \mathbb{R}^{d_m}$; \textbf{(iv)}~computes an exponential moving average ($\operatorname{ema}$) $\outclr{r_t} \in \mathbb{R}$ of the linear memory \outclr{readout} $\outclr{\mathbf{w}_r}^\top \memclr{\mathbf{m}_t}$ using timescale $\outclr{\tau_r} \in \mathbb{R}$; and \textbf{(v)}~applies a ReLU activation with a per-neuron bias to the readout with its $\operatorname{ema}$ subtracted, implementing a temporal high-pass filter, and yielding the neuron's activity $a_t \in \mathbb{R}$. Concretely, each neuron implements $\text{ELM}(\boldsymbol{z}_t)$ defined by:
\begin{equation}
\begingroup
\renewcommand{\arraystretch}{1.10}
\begin{array}{r@{\quad}r@{{}={}}l}
\text{(i)}   & \synclr{\boldsymbol{b}_t}       & c \cdot \operatorname{branch\_sum}(\boldsymbol{z}_t \odot \synclr{\boldsymbol{w}_s}) \\
\text{(ii)}  & \mlpclr{\Delta\boldsymbol{m}_t}  & \tanh\!\left(\operatorname{MLP}_{\mlpclr{\boldsymbol{w}_p}}([\synclr{\boldsymbol{b}_t},\memclr{\boldsymbol{\kappa}_m} \odot \memclr{\boldsymbol{m}_{t-1}}])\right) \\
\text{(iii)} & \memclr{\boldsymbol{m}_t}        & \memclr{\boldsymbol{\kappa}_m} \odot \memclr{\boldsymbol{m}_{t-1}} + (1-\mlpclr{\boldsymbol{\kappa}_\lambda}) \odot \mlpclr{\Delta\boldsymbol{m}_t} \\
\text{(iv)}  & \outclr{r_t}                     & \outclr{\kappa_r}\outclr{r_{t-1}} + (1-\outclr{\kappa_r})\outclr{\boldsymbol{w}_r}^{\top}\memclr{\boldsymbol{m}_t} \\
\text{(v)}   & a_t                              & \operatorname{ReLU}\!\left(b + \outclr{\boldsymbol{w}_r}^{\top}\memclr{\boldsymbol{m}_t} - \outclr{r_t}\right)
\end{array}
\endgroup
\label{eq:elm_equations}
\end{equation}
where $\outclr{\kappa_r} = \exp(\nicefrac{-1}{\outclr{\tau_r}})$,  $\memclr{\boldsymbol{\kappa_m}} = \exp(\nicefrac{-1}{\memclr{\boldsymbol{\tau_m}}})$,  $\mlpclr{\boldsymbol{\kappa_\lambda}} = \exp(\nicefrac{-\mlpclr{\lambda}}{\memclr{\boldsymbol{\tau_m}}})$. Learnable: $\synclr{\boldsymbol{w_s}}, \mlpclr{\boldsymbol{w_p}}, \outclr{\boldsymbol{w_r}}, b$. Fixed: timescales $\memclr{\boldsymbol{\tau_m}}, \outclr{\tau_r}$, input scale $c$, and timescale ratio $\mlpclr{\lambda}$. The $\text{branch\_sum}$ sums contiguous segments of size $d_\text{branch}$. The neuron's expressivity is primarily controlled through $d_m$, $l_\text{mlp}$, and $\tau_\text{max}$.

\vspace{-0.5em}
\paragraph{The ELM Layer.}
An ELM layer consists of $N_\text{rec}$ independently parameterized ELM neurons, each receiving $P/N_{\mathrm{rec}} = k_e + k_c$ trainable parameters of the layer budget $P$. The layer input at time $t$ is the concatenation of the feed-forward signal $\mathbf{u}_{t}$ (either external input or the previous layer's activity) and the layer's own previous activity $\mathbf{a}_{t-1}$. Each neuron receives a fixed, uniformly sampled subset of $d_s$ input channels from this concatenation, with $\rho_\text{rec}$ determining the probability of sampling a recurrent connection at initialization. A layer, therefore, computes:
\begin{equation}
\begin{aligned}
\boldsymbol{z}_{t,i} &= \operatorname{input\_select}_i([\boldsymbol{u}_t, \boldsymbol{a}_{t-1}]), \\
a_{t,i} &= \operatorname{ELM}_i(\boldsymbol{z}_{t,i}).
\end{aligned}
\end{equation}
where $\boldsymbol{a_t} \in \mathbb{R}^{N_\text{rec}}$, $\boldsymbol{u_t} \in \mathbb{R}^{d_\text{inp}}$. $\text{input\_select}_i$ gathers the $d_s$ channels assigned to neuron $i$ at initialization, and $\text{ELM}_i$ has neuron-$i$-specific parameters.

\vspace{-0.5em}
\paragraph{The ELM Network.}
The full model comprises a larger recurrent hidden ELM layer (e.g., $N_\text{rec}=1024$), a smaller feed-forward readout ELM layer with simpler neurons ($l_\text{mlp} = 0$, $d_m = 3$) without output filtering or activation nonlinearity, and a final linear layer that maps the readout activity to the target output dimensionality. Text input sequences are embedded via a scaled one-hot encoding.

\clearpage
\section{Related Work}

\textbf{Recurrent and modular architectures} split the computation across interacting stateful components. 
Typically, \textbf{computational neuroscience} asks how feedback and population dynamics in networks of simple spiking or rate-based neurons support computation \citep{maass1997networks,jaeger2002tutorial,buonomano2009state,mante2013context,dayan2005theoretical}. However, more intricate neuron models with some tunability have also been used in networks \citep{markram2015reconstruction,sacramento2018dendritic,brenner2022tractable,zheng2024temporal}. 
Generally, \textbf{deep learning} has focused on performance and found specialized modules  \citep{koutnik2014clockwork,santoro2018relational,goyal2021recurrent,beck2024xlstm} and parallel module-like multi-head and mixture-of-experts setups \citep{vaswani2017attention,shazeer2017outrageously} to be beneficial; recently, also structured state-spaces \citep{smith2023simplified,gu2023mamba}. While most of them use vector-valued module outputs (breaking the neuron analogy) and do not separately control module connectivity, some do feature more neuron-like scalar outputs \citep{gu2023mamba,liu2025kan,darlow2026continuous}. 
Uniquely, ELM Networks feature independently parameterized scalar-output neurons with flexibly adjustable internal complexity that is separately tunable from connectivity.

\textbf{Performance scaling laws} provide the natural language for turning architectural choices into resource-allocation questions. Prior work has shown that performance often follows predictable trends with model size, data, and compute across architectures and modalities \citep{hestness2017deep,kaplan2020scaling,henighan2020scaling,hoffmann2022training}, whereas \textbf{budget-constrained tradeoff} investigations explore how finite resources should be allocated across model-shape dimensions \citep{tan2019efficientnet,dollar2021fast,alabdulmohsin2023getting}. Similar budgeted "module" complexity vs. counts experiments in related architectures have also found interior optima \citep{vaswani2017attention,amsel2025quality,dubinin2026improved,chirkova2020deep,lobacheva2020power,ludziejewski2024scaling}.
Complementary \textbf{approximation-theoretic} work studies how expressivity changes with architectural choices such as width and depth \citep{barron1993universal,cybenko1989approximation,montufar2014number,telgarsky2016benefits,raghu2017expressive}. In contrast to previous works scaling networks around a fixed computational unit or jointly with connectivity, here, the unit complexity itself and connectivity separately become scaling axes.

\textbf{Information-theoretic perspectives} on neural computation study how neural systems represent signals under noise, redundancy, and resource constraints
\citep{shannon1948mathematical,barlow1961possible,attneave1954some,laughlin1981simple}.
Efficient coding, sparse coding, information bottleneck, and population coding provide complementary views on how signals are compressed, made task-relevant, and distributed across noisy neural ensembles
\citep{barlow1961possible,olshausen1996emergence,tishby1999information,abbott1999effect,averbeck2006neural,morenobote2014information}, with encoding tradeoffs \citep{tkacik2010optimal,doi2014simple,ganguli2014efficient}.
Our framework follows this tradition, but casts the problem as a resource-allocation tradeoff between single-unit fidelity and population-level aggregation and coordination.

\vspace{-0.5em}
\section{Results} 
\vspace{-0.5em}
\label{sec:main_experimental_section}

In our experimental section, we address three core questions: \textbf{1)} How does a recurrent layer's performance scale with neuron count, neuron complexity, and network connectivity, subject to budget constraints? Section \ref{sec:large_scale_experimenta}. \textbf{2)} Can the scaling tradeoffs be explained in terms of general information-theoretic principles? Section \ref{sec:information_theory_framework}. \textbf{3)} Can we find architectural scaling recipes that trace the Pareto frontier across datasets and parameter budget? Section \ref{sec:pareto_optimal_scaling}. For complete model hyperparameters, detailed training methods, and full derivations, see Appendix \ref{sec:architecture_training_dataset} and \ref{sec:eri-derivation}.
\vspace{-0.5em}

\subsection{Evaluating network scaling and tradeoffs across a wide range on complementary datasets}
\label{sec:large_scale_experimenta}

Our first set of experiments is conducted on the SHD-Adding benchmark \citep{spieler2024the}, a challenging neuromorphic sequence benchmark with biologically plausible spatio-temporal input patterns. Each input sample is a $700 \times 1000$ binary spatio-temporal pattern, representing the concatenation of two spoken digits in German or English (see Fig.~\ref{fig:shd_data_visualization}). Individual digits were originally spike-encoded using a bio-inspired cochlear model \citep{cramer2020heidelberg}. The prediction target is the sum of the spoken digits.

\begin{figure*}[ht]
    \centering
    \includegraphics[width=0.95\linewidth]{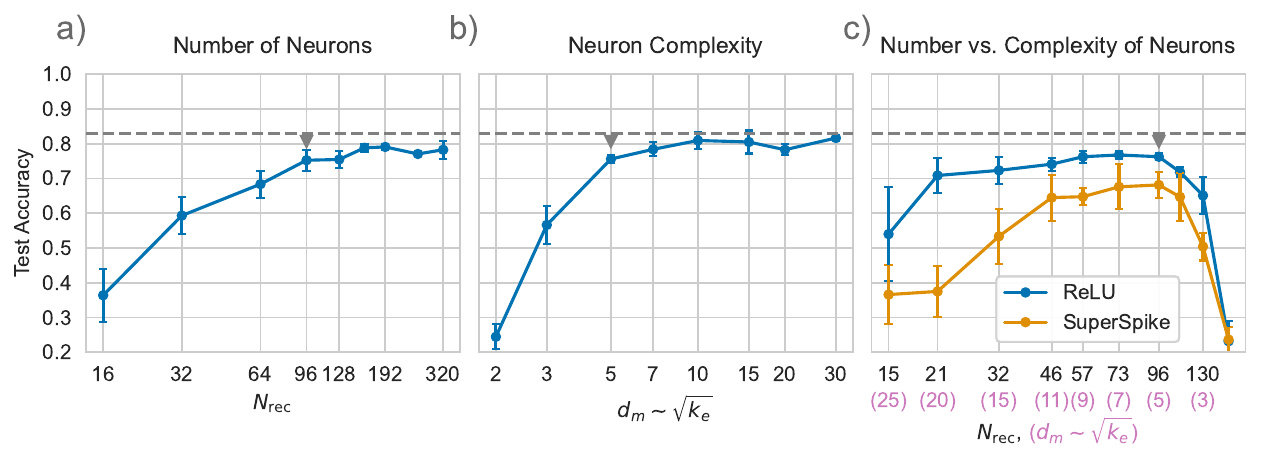}
    \vspace{-1em}
    \caption{\textbf{More neurons or more complex neurons each improve performance on the SHD-Adding task; under a fixed budget, a clear non-trivial optimum emerges.}
    The reference model (triangle) was chosen to be below saturation along all dimensions. Mean and standard deviation across three seeds are displayed.
    Test accuracy improves with \textbf{a)} the number of neurons $N_{\mathrm{rec}}$ and \textbf{b)} neuron complexity $k_e \sim d_m^2$ (varying memory units $d_m$).
    \textbf{c)} Under a fixed parameter budget, a clear optimum emerges in the number-vs-complexity tradeoff; the optimum sharpens with binary-spiking neuron activation SuperSpike \citep{zenke2018superspike} (default ReLU). Secondary $d_m$ axis label in pink brackets.
    }
    \label{fig:shd_empirical_results}
    \vspace{-1em}
\end{figure*}

\vspace{-0.5em}
\paragraph{Should one invest parameters in network width or neuron complexity?}
Through robust initialization and stable training, ELM networks manage to reliably solve the task, and we observe roughly monotonic performance improvements for scaling along the number of neurons or neuron complexity (Fig.~\ref{fig:shd_empirical_results}a,b),  with networks of roughly 150 neurons or 10 memory units approaching the best previously reported accuracy levels ($83\%$ \citep{spieler2024the}). For a fixed parameter budget, however, these two dimensions compete, and a non-monotonic curve emerges (Fig.~\ref{fig:shd_empirical_results}c), with a steep drop-off in performance for networks with small neurons or few neurons, and a clear in-between optimum that sharpens for neurons trained with binary activation as in spiking neural networks. Thus, the parameter budget in recurrent layers should be carefully balanced between complexity and the number of neurons.

Despite the dataset being one of the most challenging neuromorphic benchmarks, even modest-sized ELM Networks start running into dataset-related saturation effects. To explore scaling across larger network dimensions in all directions, we turn to a complementary character-level language modeling benchmark, Enwik8 \citep{mahoney2011large}. It is comprised of the first $10^8$ bytes from Wikipedia in 2006, containing 240k articles primarily in English. By using the standard preprocessing pipeline from Transformer XL \citep{dai2019transformer}, we get 204 unique bytes, which we one-hot encode as the network input.

\begin{figure*}[ht]
    \centering
    \includegraphics[width=0.93\linewidth]{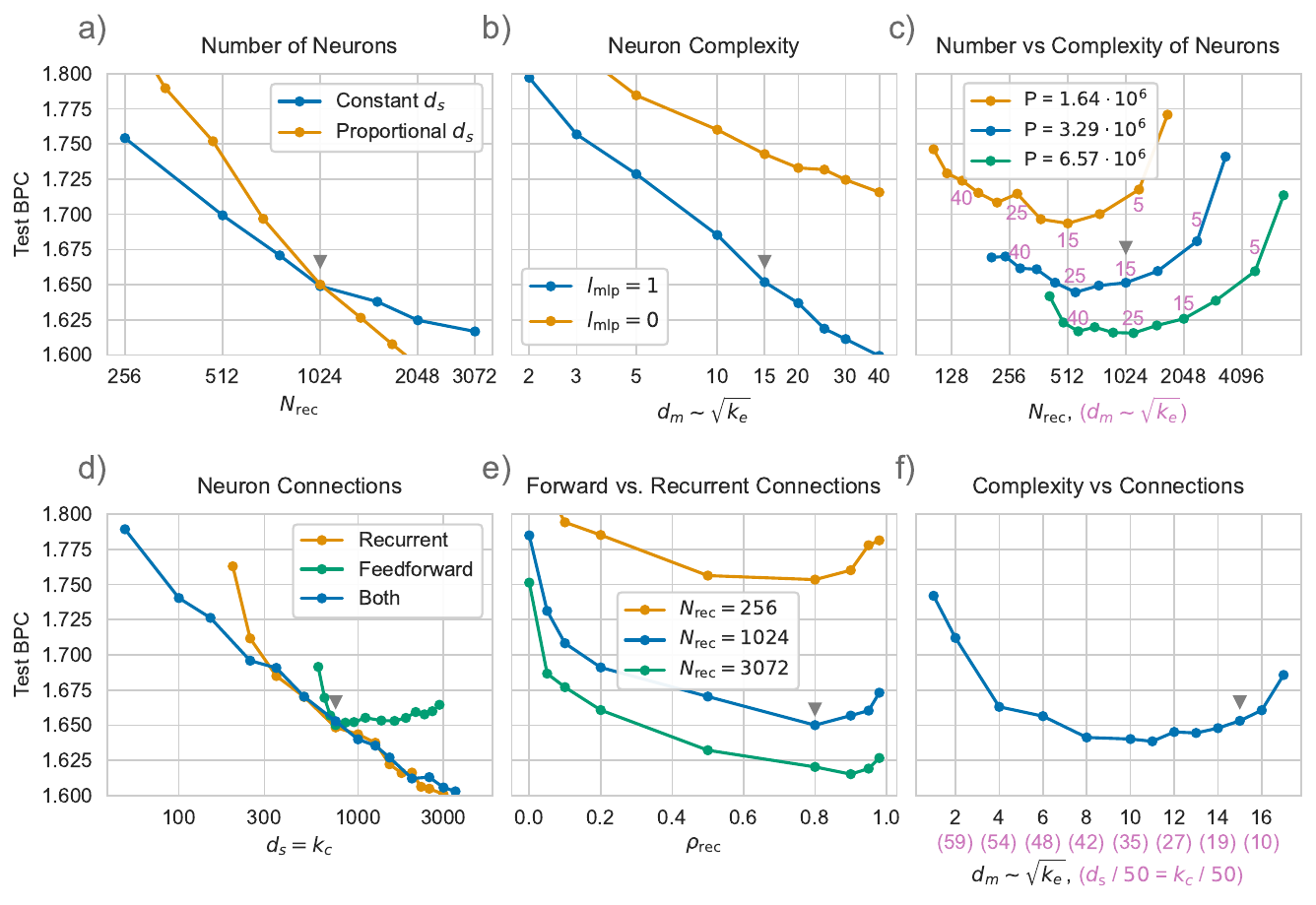}
    \vspace{-1em}
    \caption{\textbf{Monotonic scaling gains extend across vast network sizes; larger budgets favor both more \emph{and} more complex neurons, and connectivity introduces new tradeoff dimensions.}
    Enwik8: character-level language modeling (test BPC, lower is better) \citep{mahoney2011large}. Reference model marked with a triangle. Exemplary network activity in Appendix Fig.~\ref{fig:enwik8_network_activity_visualization}.
    \textbf{a, b)} Monotonic improvement with neuron count $N_{\mathrm{rec}}$ and per-neuron complexity $k_e \sim d_m^2$ (varying memory units $d_m$) spanning over an order of magnitude each without saturating, with slope of improvement modulated by network connectivity and neuron integration hierarchy respectively.
    \textbf{c)} Under a fixed budget, a clear non-trivial optimum emerges. It shifts with increased budget, favoring more complex and more numerous neurons. Similar tradeoffs are observable in multi-head attention Transformers; see Appendix Fig.~\ref{fig:transformer_head_tradeoff}.
    \textbf{d)} More synapses $d_s$ improve performance, and recurrent connections drive this improvement far beyond where feedforward connections saturate.
    \textbf{e)} An optimal recurrent fraction $\rho_{\mathrm{rec}}$ shifts with network size, roughly tracking the ratio of recurrent neurons to possible presynaptic inputs.
    \textbf{f)} Splitting neuron parameters between complexity and connectivity yields an optimum around $N_\mathrm{rec}$ connections.}
    \label{fig:enwik8_empirical_results}
    \vspace{-1em}
\end{figure*}

\paragraph{Do these scaling tradeoffs generalize to larger dataset and network sizes?}
The monotonic performance improvement trends observed on SHD-Adding extend on Enwik8 over at least another order of network width and neuron complexity (Fig. \ref{fig:enwik8_empirical_results}a,b). Curiously, the individual scaling slopes can be manipulated: by increasing neuron input connections ($d_s$) proportional to neuron count, the network improves much faster, and if we impose simpler synaptic integration ($l_{\mathrm{mlp}}=0$), we see slower performance improvements, even when accounting for total parameters (see Appendix Fig.~\ref{fig:enwik8_neuron_network_scaling_params_matched}). Lastly, we again find a nontrivial optimal neuron complexity for a resource-constrained layer, with the optimum emerging robustly across three parameter budgets (Fig.~\ref{fig:enwik8_empirical_results}c). Interestingly, we find that increased budget favors network configurations with more \emph{and} more complex neurons; shifting from $(N_\mathrm{rec},d_m)=(512,15)$ to $(1024,25)$ when the layer budget is quadrupled. Importantly, this interior-optimum emerges robustly also for different levels of network recurrence or number of network hidden layers (Appendix Fig. \ref{fig:enwik8_scaling_in_ff_and_2hidden_networks}), wide range of neuron high-pass-filter timescales (Appendix Fig.\ref{fig:enwik8_readout_tau_sensitivity_analysis}), and even alternative budget types like Wall-Time or FLOPS (Appendix Fig. \ref{fig:iso_budget_ablations}).

\paragraph{How does network performance scale with connectivity and recurrence?}
Treating connectivity as its own scaling dimension, we observe similar monotonic performance improvements (Fig.~\ref{fig:enwik8_empirical_results}d). However, if we change the number of recurrent or feedforward connections individually, a more nuanced picture emerges: while having too few of either results in a stark drop-off in performance, once sufficient feedforward connections from input exist, almost all performance benefit comes from the additional recurrent connections, a trend that continues well into the regime of multiple connections between pairs of neurons ($d_s \gtrsim N_\mathrm{rec} + d_\mathrm{inp}=1228$).
When varying the fraction of recurrent connections to total inputs, $\rho_\mathrm{rec}$, we find that larger networks favor higher recurrence (Fig.~\ref{fig:enwik8_empirical_results}e).
Lastly, keeping the number of neurons fixed but varying the split of parameters between connectivity $k_c$ and neuron complexity $k_e$, we find an optimum around $N_\mathrm{rec}$ connections (Fig.~\ref{fig:enwik8_empirical_results}f). 


\subsection{Evaluating network scaling and tradeoffs under an information-theoretic framework}
\label{sec:information_theory_framework}

Our previous experiments demonstrate the advantage of neurons with intermediate complexity empirically, yet also raise new questions: why is it better to invest additional parameter budget simultaneously in more \emph{and} more complex neurons, or how exactly do neuron integration capabilities and network connectivity modulate scaling steepness?
To get a better understanding of the underlying dependencies, we develop a theoretical framework and establish its link to experimental observations.

\begin{figure*}[ht]
    \centering
    \includegraphics[width=1.0\linewidth]{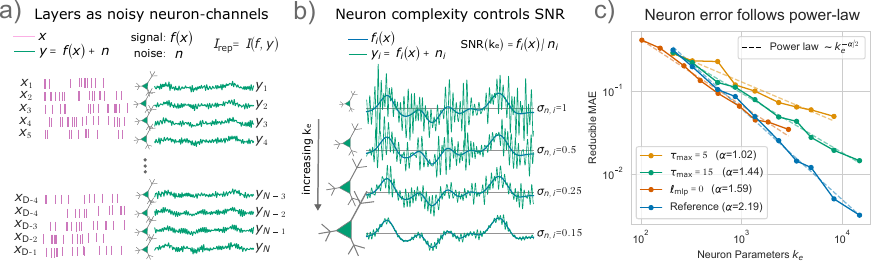}
    \vspace{-0.75em}
    \caption{\textbf{The \emph{Effective Representation Information} derived from viewing neural layers as noisy channels is motivated by residual-error scaling in single neurons.}
    \textbf{a)} Neural activations may be viewed as a noisy multi-channel representation of their inputs, with  $I_{\mathrm{rep}}$ quantifying how much task-relevant information they transmit.
    \textbf{b)} Per-neuron complexity determines the residual variance $\sigma_n^2(k_e)$, a decreasing function of parameter budget $k_e$, that leads to increasing signal-to-noise ratio $\mathrm{SNR}(k_e)$. 
    \textbf{c)} We test this noise ansatz empirically by fitting increasingly complex ELM neurons to single-neuron membrane-voltage recordings from the NeuronIO dataset \citep{beniaguev2021single}: the average (three seeds) reducible Mean Absolute Error (MAE) on test set decreases as a power law in neuron parameters $\propto k_e^{-\alpha/2}$ across two orders of magnitude, with $\alpha$ capturing the neuron's spatio-temporal integration settings (e.g.\ $l_{\mathrm{mlp}}$, $\tau_{\mathrm{max}}$).
    This new perspective on neural computation motivates the particular form of $I_\mathrm{rep}$.
    }
    \label{fig:theory_introductory_figure}
    \vspace{-1em}
\end{figure*}

\paragraph{An information-theoretic model of a neural layer.}
We frame a neural layer as a collection of noisy neuron-channels (Fig.~\ref{fig:theory_introductory_figure}a,b): each neuron $i$ produces an output $y_i = f_i(x) + n_i$, where $f_i$ is a task-relevant signal component, and $n_i$ is a task-uninformative residual (approximation error), treated as phenomenological noise \citep{bialek1989reading,borst1999information}. 
Consequently, the latent signal component $f_i(x)$ is not directly observable, as measurements of a neuron's output $y_i$ will also contain its task-uninformative noise component $n_i$.
Unlike previous works, we further model per-neuron complexity $k_e$ reducing its residual noise (approximation error) $n_i$, making each neuron-channel output closer to its ideal $f_i$. 
The mutual information $I_{\mathrm{rep}} = I(f; y)$ between a layer's noisy output and its task-relevant signal, the \emph{effective representation information}, quantifies how much usable information the layer carries, and is therefore a proxy indicative of task performance.
More channels \emph{or} more informative neuron channels increase $I_{\mathrm{rep}}$, inter-neuron signal redundancy reduces it relative to independent channels.

Four assumptions together yield a closed-form expression for $I_{\mathrm{rep}}$: \textbf{(A1)} Gaussian signal and noise; \textbf{(A2)} power-law signal covariance eigenvalue spectrum $\lambda_i \propto i^{-\beta}$, with smaller $\beta$ corresponding to a higher effective dimensionality of the signal across channels; \textbf{(A3)} per-neuron noise decays as a power law in $k_e$ until saturating at a floor $q_\infty$, $\sigma_n^2(k_e) \propto \max((\gamma k_e)^{-\alpha}, q_\infty)$, and \textbf{(A4)} a fully fungible budget $P$, distributed among $N = P/(k_e+k_c)$ neurons each receiving $k_e$ effective parameters under connectivity overhead $k_c$, concretely
\begin{equation}\label{eq:I_rep}
  I_{\mathrm{rep}}(k_e) \;=\; \tfrac{1}{2}\sum_{i=1}^{P/(k_e+k_c)} \log_2\!\left(1 + s(k_e)\, i^{-\beta}\right), \qquad s(k_e) \;=\; \min\!\left((\gamma k_e)^\alpha, q_\infty^{-1}\right).
\end{equation}
with $s(k_e)$ as the neuron's parameter dependent signal-to-noise ratio (SNR). The budget trade-off is visible directly: larger $k_e$ raises per-channel SNR but reduces the number of channels $N$, creating a tension and raising an allocation question.
Full derivation in Appendix~\ref{sec:eri-derivation}.

\begin{figure*}[ht]
    \centering
    \includegraphics[width=1.0\linewidth]{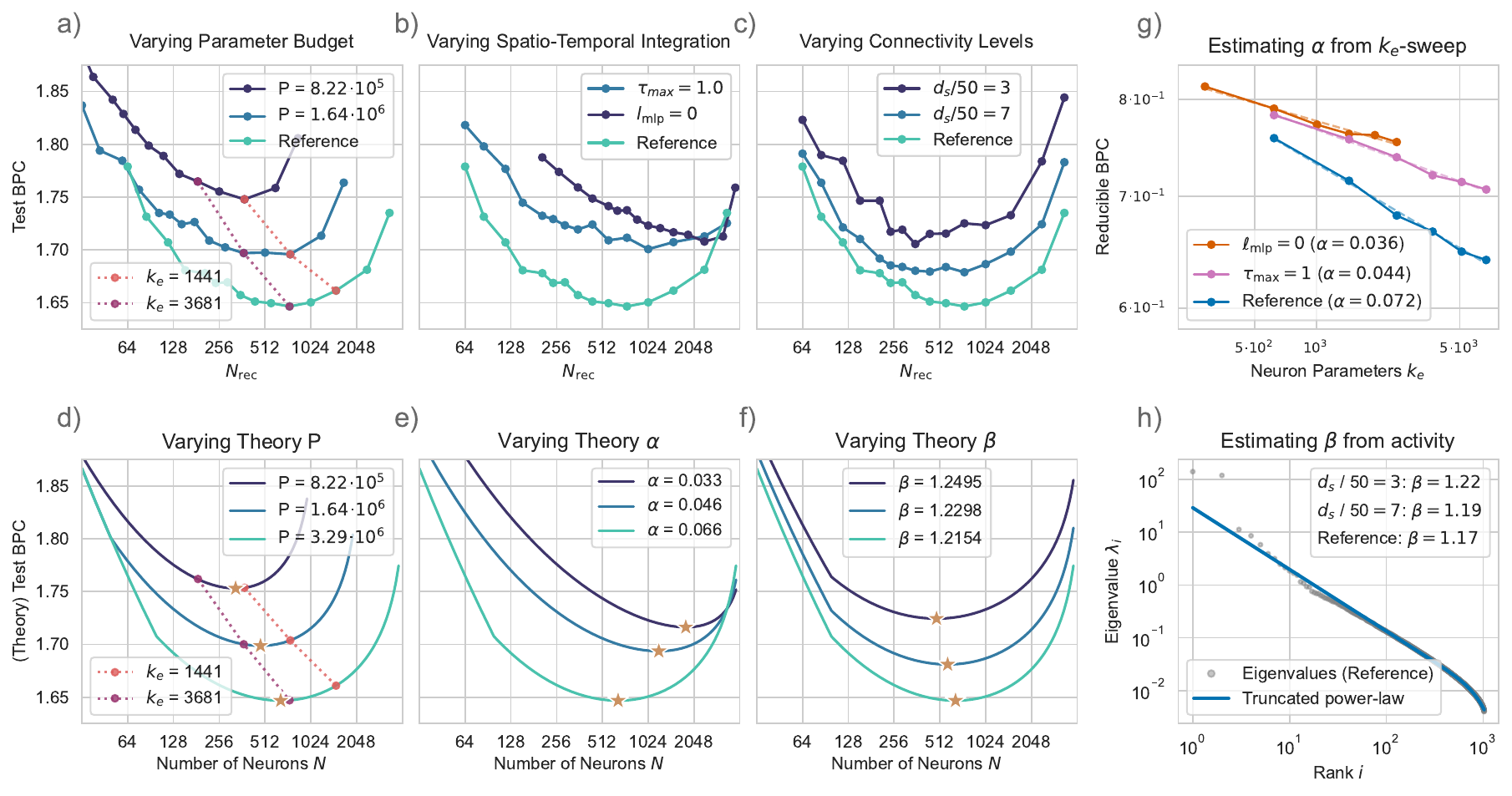}
    \vspace{-1em}
    \caption{\textbf{The theoretical model qualitatively reproduces the empirical scaling tradeoffs and links their shifts to measurable architectural quantities.}
    \textbf{a--f)} Enwik8 experiments (top row) are paired with corresponding theory proxy (bottom row) obtained via joint fit using shared reference parameters (Pearson $r = 0.9828$ and RMSE $= 0.0092$ BPC). In all panels, the reference model has $d_s = 750$ (synapse connections), $\tau_\mathrm{max}=100$ (longest memory timescale),$l_\mathrm{mlp}=1$ (MLP hidden layers), and theory is affine adjusted using $\mathrm{BPC} = -0.34\cdot I_{\mathrm{rep}} + 2.7$. Theory optima marked with a golden star. For panels a--f overlaid see Appendix Fig. \ref{fig:theory_scaling_overlay}.
    \textbf{a, d)}  Optimal performance for intermediate neuron complexity; increasing the budget $P$ shifts the optimum toward more and larger neurons. Curves touch for large $k_e$ once the neurons reach the per-neuron noise floor $q_\infty$.
    \textbf{b, e)} Weakening spatio-temporal integration matches decreasing $\alpha$ and shifts the optimum toward wider layers. All curves cross at a signal-to-noise ratio of one with $k_e=1/\gamma$.
    \textbf{c, f)} Reducing connectivity corresponds to increasing the spectral decay exponent $\beta$, and shifts the optimum toward larger neurons.
    \textbf{g, h)} Direct measurements of $\alpha$ and $\beta$ on the models with $N_\mathrm{rec}=1024$ are in the same ballpark \emph{and} change in accord with the joint theoretical fits' values.
    \textbf{g)} Estimating $\alpha$ by fitting a power law to the dependence of reducible BPC on single neuron parameter  (details Appendix \ref{sec:irep-scaling}).
    \textbf{h)} Estimating $\beta$ from the eigenvalue spectrum of the network activity covariance matrix (at memory readout $\boldsymbol{w_r}^\top \boldsymbol{m_t}$) by fitting a truncated power law with shared cutoff parameters $(i_c, \nu)$ (details in Appendix Fig.~\ref{fig:enwik8_eigenvalue_spectrum}).
    For \textbf{predictive framework} experiments, with all theory parameters independently derived, and the proxy predicting budgeted tradeoffs (Pearson $r = 0.9801$), see Appendix Fig. \ref{fig:theory_predictions}}
    \label{fig:theory_scaling}
    \vspace{-1em}
\end{figure*}

\paragraph{What do the theory parameters mean in ELM Networks, and do the assumptions hold?}
While some theory parameters are directly accessible from the architecture, the four phenomenological parameters require some deliberation and rest on assumptions that we will examine in the following. 
\textbf{The accessible ones:} $N = N_\mathrm{rec}$ is the recurrent-layer width, $k_e = \#w_p + \#w_r$ is the per-neuron internal parameter count, and $k_c = \#w_s$ is the per-neuron connection parameter count.
\textbf{The noise decay exponent} $\alpha$ describes an individual neuron's spatio-temporal integration expressivity, which also depends on $\tau_\mathrm{max}$ or $l_\mathrm{mlp}$. The underlying assumption (A3) is testable on the NeuronIO single-neuron benchmark, where per-neuron noise is directly observable as the residual error in fitting an isolated ELM neuron to recorded membrane voltages: the reducible MAE follows a power law in $k_e$ across two orders of magnitude, yielding $\alpha$ from the slope (Fig.~\ref{fig:theory_introductory_figure}c). Within ELM Network trained on Enwik8, the per-neuron noise reductions remain close to power law, with its slope tracking the neuron-related expressivity changes qualitatively (Fig.~\ref{fig:theory_scaling}g).
\textbf{The spectral decay exponent} $\beta$ captures the covariance structure of the task-relevant signal component across neurons (smaller $\beta$, higher effective dimensionality, less inter-neuron redundancy), which we expect to be shaped by connectivity: more connections allow neurons to coordinate, reducing redundancy and decreasing $\beta$. We test (A2) on ELM Networks, as unlike $\alpha$, $\beta$ is a neural layer property and cannot be investigated in individual neurons: the empirical eigenvalue spectrum of the layer's activity covariance matrix is well approximated by a truncated power law with a low-rank cutoff (Fig.~\ref{fig:theory_scaling}h, Appendix Fig.~\ref{fig:enwik8_eigenvalue_spectrum}). (A2) is thus broadly supported in our setting, with deviations confined to the low-eigenvalue tail that contributes negligibly to $I_\mathrm{rep}$.
\textbf{The remaining parameters}, $\gamma$ and $q_\infty$, set the prefactor and floor of (A3): $\gamma$ converts model parameter count into dimensionless effective parameters, a conversion efficacy that may be influenced by train steps or optimizer settings, while $q_\infty$ is the residual variance no amount of $k_e$ can reduce, observable as saturation in the error-vs-$k_e$ curve (Appendix Fig.~\ref{fig:neuronio_powerlaw},\ref{fig:theory_qinf_comparison}).
Note that these parameters are generally \textbf{model- and dataset-specific}; for example, $\alpha$ directly captures neuron-model-specific expressivity, and $\beta$ also depends on the underlying training data \citep{maloney2022solvable}.

\paragraph{Do model predictions match the experimental scaling curves?}
We probe the theoretical framework and its parameter-to-architecture associations through matched theory-to-experiment comparisons on Enwik8 (Fig.~\ref{fig:theory_scaling}). The experimental evaluations are paired with theoretical curves obtained from a single joint fit: the curves of Fig.~\ref{fig:theory_scaling}a/d and reference curves in (b/e, c/f) share a single tuple $(\alpha, \beta, \gamma, q_\infty)$ and affine mapping $I_\mathrm{rep}\to\mathrm{BPC}$. Each set of experiments targets one of the proposed architectural-to-theory correspondences: budget $P$, the noise-decay exponent $\alpha$, or the spectral exponent $\beta$, all other parameters are kept at reference levels.

We find: \textbf{(1)} the joint fit qualitatively reproduces the empirical scaling shapes for all three sensitivity analysis groups, including the location and shift of the optimum with budget (a/d), the wider optima and curve crossings under reduced $\alpha$ (b/e), and the vertical shifts and slight optimum shifts under reduced $\beta$ (c/f); \textbf{(2)} the fitted $\alpha$ and $\beta$ values are in the same ballpark as independent in-network measurements (Fig.~\ref{fig:theory_scaling}g,h), with $\alpha$ biased slightly high and $\beta$ uniformly biased slightly low, compatible with the $\alpha$/$\beta$ fitting degeneracy; and \textbf{(3)} the response of measured $\alpha$ and $\beta$ to architectural changes agrees in direction with the joint-fit predictions, $\alpha$ decreasing when neuron integration is weakened ($\tau_\mathrm{max}=1$, $l_\mathrm{mlp}=0$) and $\beta$ increasing when connectivity is reduced ($d_s$ lowered).
Together, a four-parameter proxy captures the empirical scaling tradeoffs across three architectural axes under a single constrained fit, with each phenomenological exponent independently identifiable from architecture-specific measurements. Note that $\alpha, \beta, \gamma$ and $q_\infty$ are defined architecture-agnostically, and may be applicable to any layer of scalar-output units with separately tunable complexity.

\subsection{Search for optimal architecture scaling rule via large-scale hyperparameter exploration}
\label{sec:pareto_optimal_scaling}

The theoretical proxy can predict the optimum's location in $(N_\mathrm{rec}, k_e)$, but treats $k_e$ and $k_c$ as a scalar parameter count, leaving the internal parameter allocation across $d_m$ (memory units), $d_\mathrm{mlp}$ (MLP width), $d_\mathrm{tree}$ (number of branches), $d_\mathrm{branch}$ (synapses per branch) unspecified. We address this with a large-scale, structured search, over $(N_{\mathrm{rec}}, d_m)$ together with configuration rules that constrain $d_{\mathrm{mlp}}, d_{\mathrm{tree}},$ and $d_{\mathrm{branch}}$ subject to $d_m \leq d_{\mathrm{mlp}} \leq d_{\mathrm{tree}} \leq d_{\mathrm{branch}}$, and set the recurrence fraction $\rho_{\mathrm{rec}} \sim \nicefrac{N_{\mathrm{rec}}}{(N_{\mathrm{rec}}+d_{\mathrm{inp}})}$ for both SHD-Adding and Enwik8 tasks (Fig.~\ref{fig:adaptive_scaling}). Details in Appendix \ref{sec:structured_joint_hp_search_method_details}.

\begin{figure*}[ht]
    \centering
    \includegraphics[width=1.0\linewidth] {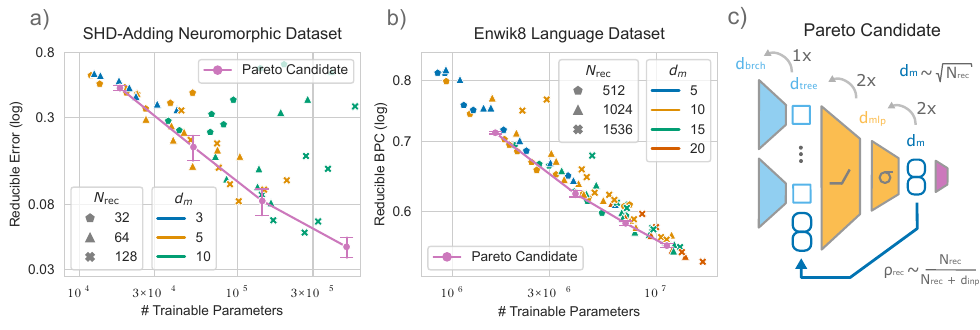}
    \vspace{-1em}
    \caption{\textbf{A simple joint-scaling heuristic consistent with theory traces the empirical Pareto frontier across both datasets and three orders of magnitude.}
    We perform a structured large-scale hyperparameter search including neuron internal dimensions. The mean across three runs is reported on SHD-Adding and single-run performance on Enwik8. For the Pareto candidates, the mean and standard deviation across five runs are shown for both datasets.
    \textbf{a,b)} On both datasets, networks with more \emph{and} more complex neurons become optimal as budget increases (as before), and a single joint scaling recipe displaying approximate power-law decay can trace the Pareto frontier closely.
    \textbf{c)} The \emph{Pareto Candidate} only has  $N_{\mathrm{rec}}$ as free scaling variable with all other settings derived from it.}
    \label{fig:adaptive_scaling}
    \vspace{-1.0em}
\end{figure*}

We find a single scaling recipe to perform routinely among the best across both datasets and roughly three orders of magnitude in trainable parameters: $d_m \sim \sqrt{N_{\mathrm{rec}}}$ with $d_{\mathrm{mlp}} = 2d_m$, $d_{\mathrm{tree}} = 2d_{\mathrm{mlp}}$, and $d_{\mathrm{branch}} = d_{\mathrm{tree}}$ 
scaling per-neuron parameters and connectivity linearly with $N_{\mathrm{rec}}$. 
The recipe scales per-neuron connectivity aggressively, in line with the theoretical proxies corresponding $\beta$'s conceptual importance; the resulting reducible-error decay follows an approximate power law consistent with above dataset and neuron noise floor regimes.
The theoretical proxy predicts the recipe to hold only until larger neurons reach their individual noise floor $q_\infty$, after which any additional benefit must come from more and better coordinated neurons; a regime we have not yet reached in this experiment  ($d_m \leq 20$ vs $d_m \approx 70$ on Enwik8), and which we leave for future work.

\section{Discussion}
\label{sec:discussion_limitations}

In this work, we studied the empirical and theoretical performance tradeoff between the number, complexity, and connectivity of neurons in recurrent layers under budget constraints. Empirically, performance improves monotonically along each axis, but under a fixed budget, a non-trivial optimum emerges that shifts toward both more \emph{and} more complex neurons as the budget grows. In our information-theoretic framework, the dependence of \emph{Effective Representation Information} ($I_{\mathrm{rep}}$) on the number and complexity of channels reproduces these tradeoff shapes and their shifts under a single constrained joint fit, with each phenomenological exponent independently identifiable from architecture-specific measurements. A joint search yields a simple scaling recipe for model hyperparameters that traces the empirical Pareto frontier across both studied datasets and three orders of magnitude in trainable parameters, consistent with the theoretical proxies' predictions. Our findings position complex per-unit machinery as a budget-efficient design choice, with implications for both architecture design in machine learning and the interpretation of biological neuron complexity.

\paragraph{Limitations:} 
All results are for ELMs and ELM Networks, limiting the generality of the findings.
\textbf{Experiments.} Models with the same number of parameters can vary significantly in GPU wall-time or memory consumption, due to hardware misalignment. Using a fixed training setup likely degrades performance non-uniformly across tested configurations. Both bias the exact scaling exponents and location of optima. While extensive, the structured Pareto search may have missed optimal configurations outside its search space.
\textbf{Theory.} 
The framework abstracts away recurrence, network topology, and optimization dynamics. Its assumptions hold approximately rather than exactly (e.g. truncated vs pure power-law eigenvalue spectrum), and the architecture-to-theory mapping is empirical rather than derived. While independently measured $\alpha$ and $\beta$ yield qualitatively good proxy predictions, their exact values are not readily identifiable due to significant variance.
\textbf{Neuroscience.} The phenomenological nature of the ELM neuron, BPTT-based training, and the use of parameter count as a budget proxy that does not directly capture metabolic cost, together limit direct biological interpretability.

Despite these caveats, our results suggest that per-unit complexity and connectivity machinery deserves systematic study as a scaling axis in its own right, with the compact Effective Representation Information formulation potentially helping to explore it effectively.

Beyond our preliminary experiments with multi-head attention Transformers, signatures of interior optima for similar budgeted module count versus module size experiments in multiple other related architectures \citep{amsel2025quality,dubinin2026improved,chirkova2020deep,lobacheva2020power,ludziejewski2024scaling} indicate that the presented framework might extend to, and the observed tradeoff may be present in, more mainstream architectures.

Natural next steps include probing larger network sizes to test the predicted noise-floor regime of the scaling recipe; treating training and data as explicit scaling dimensions; extending the framework to related modular architectures and structured connectivity; and finally, providing guidance for analogous measurements in biological circuits.

\acksection{
This work was funded by the Federal Ministry of Research, Technology and Space (BMFTR) through the RealmNN project under the NeuroAI initiative (grant number 01GQ2508).
We acknowledge support from the BMBF through the T\"ubingen AI Center (FKZ: 01IS18039B).
AL is a member of the Cluster of Excellence ``Machine Learning: New Perspectives for Science'', EXC 2064/1 (project number 39072764).

We thank the Max Planck Computing and Data Facility (MPCDF) for providing computational resources on the Raven high-performance computing system.
AS thanks David Beniaguev and Antonio Orvieto for insightful discussions.

The authors declare no competing interests.
}

\newpage
\clearpage

{
\small
\bibliographystyle{unsrtnat}
\bibliography{references}
}

\newpage
\clearpage

\appendix

\section{Architecture, Training, Dataset and Analysis Details}
\label{sec:architecture_training_dataset}

\textbf{Software libraries and packages.} All numerical experiments were performed in Python 3.12.10 using JAX 0.4.35 with GPU-accelerated JAXlib 0.4.35, leveraging Equinox 0.11.10 for model construction, Optax 0.2.3 for optimization, and PyTorch 2.5.1 for data loading. Statistical analysis and curve fitting used NumPy 2.2.6 and SciPy 1.14.1. The accompanying code repository can be found at \url{https://github.com/AaronSpieler/elmnetwork}.

\textbf{Compute infrastructure and runtimes.} Accelerated computations were performed on up to four A100 40GB GPUs on shared compute clusters, primarily the Raven cluster at MPCDF, with individual runs taking no longer than 24h, typically taking 2-4h on SHD-Adding, 5-15h on Enwik8, and 8-12h on NeuronIO on a single GPU. Scaling the reference ELM Network on Enwik8 beyond $d_m=35$, $N_\mathrm{rec}=3072$ or $d_\mathrm{branch}=50$ required more than one GPU to fit in VRAM and ran longer. Approximately 75k GPU hours were spent throughout the project, a vast majority of which was spent developing the architecture ($\sim50\%$) and on exploratory or scaling experiments ($\sim30\%$).

\begin{table}[h]
\centering
\caption{Reference model configurations for all three benchmarks.}
\label{tab:reference_config}
\begin{tabular}{llccc}
\toprule
Parameter & Notation & SHD-Adding & Enwik8 & NeuronIO \\
\midrule
\multicolumn{5}{l}{\textit{Neuron}} \\
\midrule
\quad Memory units & $d_m$ & 5 & 15 & 20 \\
\quad MLP hidden layers & $l_\text{mlp}$ & 1 & 1 & 1 \\
\quad MLP hidden width & $d_\text{mlp}$ & $2d_m$ & $2d_m$ & $2d_m$ \\
\quad Number of branches & $d_\text{tree}$ & 30 & 50 & 45 \\
\quad Synapses per branch & $d_\text{branch}$ & 10 & 15 & 100 \\
\quad Total synapses & $d_s$ & 300 & 750 & 4500 \\
\quad Input scale & $c$ & 10 & 10 & 1 \\
\quad Timescale ratio & $\lambda$ & 5 & 5 & 5 \\
\quad Memory timescales & $\tau_\mathrm{min},\, \tau_\mathrm{max}$ & 1, 500 & 0.1, 100 & 1, 150 \\
\quad High-pass filter & $\tau_r$ & 5 & 2 & --- \\
\midrule
\multicolumn{5}{l}{\textit{Layer}} \\
\midrule
\quad Recurrent neurons & $N_\text{rec}$ & 96 & 1024 & --- \\
\quad Recurrence fraction & $\rho_\text{rec}$ & 0.25 & 0.8 & --- \\
\quad Readout neurons & --- & 19 & 204 & --- \\
\quad Input embedding & --- & --- & one-hot & --- \\
\bottomrule
\end{tabular}
\label{tab:param_table}
\end{table}

\paragraph{Modifications to the ELM neuron in \citep{spieler2024the}:} this work builds on the ``Branch-ELM'' variant with the improved memory update, with fixed memory timescales $\tau_m$ initialized equidistant in log-space, and disabled synapse current decay using $\tau_s=0$. The implementation adds a high pass filter on the memory readout with EMA timescale $\tau_r$ for robust and stable training, uses a ReLU neuron output activation with learnable bias $b$, scales synaptic inputs using $c$ which accelerates early training, uses default weight initialization per branch for $w_s$ too, swaps $ReLU^2$ in for MLP hidden layer activation for performance, and adds a small L2 regularization term on the MLP output for improved trainability.

\subsection{Robustness of recurrent networks with high-pass filtered neurons}

While successfully training ELM Networks to similar performance without a high-pass filter mechanism \emph{is possible}, its addition to ELM neurons is sufficient to make the initialization and training of this doubly-recurrent architecture remarkably robust with respect to orders of magnitude of parameter changes in neuron complexity and network layer width, and results in stable dynamics throughout.

Typically, such neuroscience-inspired architectures are highly sensitive to training and initialization, struggling with exploding or vanishing activity, requiring careful initialization and regularization \citep{rossbroich2022fluctuation}. Yet with this simple high-pass filter modification, standard deep learning weight initialization becomes viable, and even grossly misconfigured architectures may become simple to train.

\begin{figure*}[ht]
    \centering
    \includegraphics[width=1.0\linewidth]{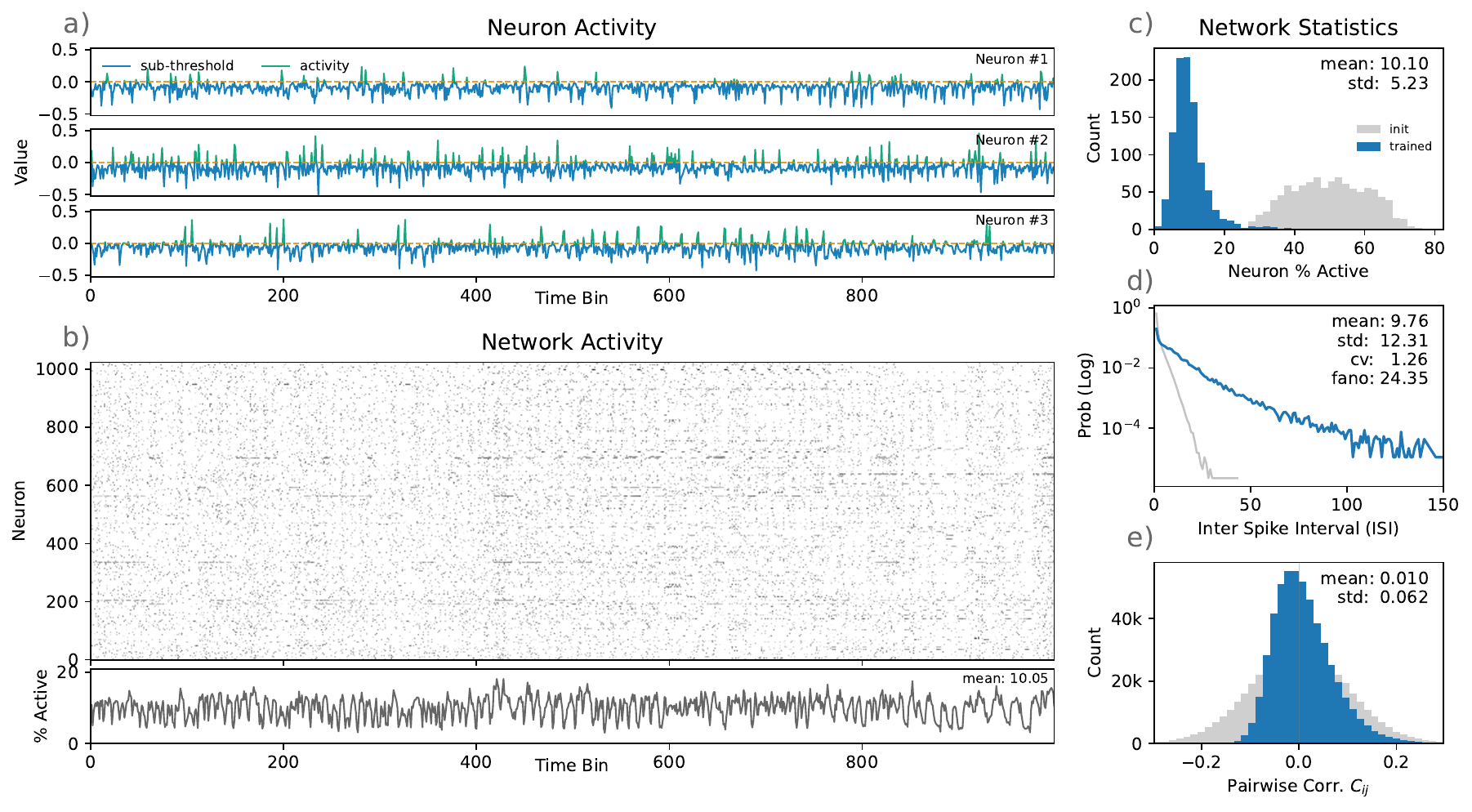}
    \vspace{-0.75em}
    \caption{\textbf{Training drives ELM Networks into a sparse, mostly asynchronous, irregular activity regime.}
    Example inference on Enwik8 of a reference model configuration with $N=1024$ and $d_m=15$.
    \textbf{a, b)} Individual neurons' activity is characterized by brief spike-like above-threshold activations and high-frequency sub-threshold fluctuations. At the population level, $\sim$10\% of neurons are active at any given time, firing asynchronously with loose global synchronization and oscillations.
    \textbf{c)} Training sparsifies the population activity from $\sim$50\% to $\sim$10\% active (with heavy tail); while a small L1 activity regularization accentuates this trend, sparsification happens regardless.
    \textbf{d)} Inter-spike intervals are approximately exponentially distributed (CV $\approx 1.3$, Fano $\approx 24$), indicating irregular bursty firing.
    \textbf{e)} Pairwise correlations are tightly centered near zero, and get smaller with training, indicating training decorrelating neuron activations.}
    \label{fig:enwik8_network_activity_visualization}
\end{figure*}

\paragraph{There are two core mechanisms why it works,} one addressing exploding activity, and one addressing vanishing activity; \textbf{1)} with sufficiently small $\tau_r$ the slow oscillations that typically blow up network activity through recurrence are completely eliminated by construction (only high frequency oscillations pass the filter), \textbf{2)} neuron and network biases that would typically keep activity levels below threshold are removed entirely, as the filtering will always result in half the signal above 0 as long as the hidden state shows \emph{any} variance at all, which translates into 50\% network activity with a 0 activity-threshold initialization by construction (any bias removed through centering) (see Fig.~\ref{fig:enwik8_network_activity_visualization}).

\begin{figure*}[ht]
   \centering
   \includegraphics[width=0.8\linewidth]{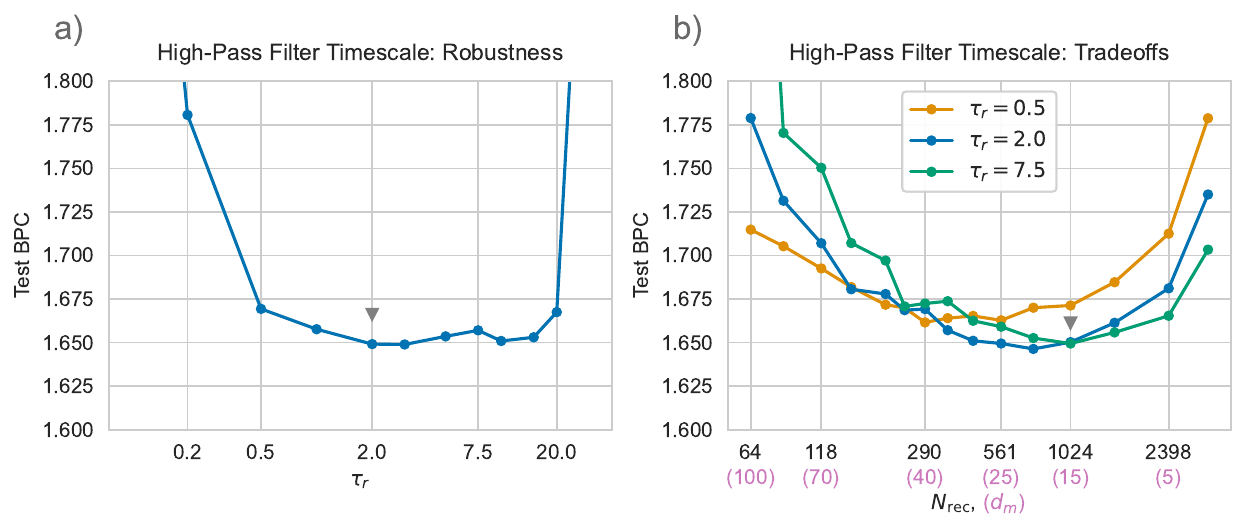}
   \caption{\textbf{A wide range of high-pass filter timescales stabilize network dynamics and result in the same qualitative tradeoffs:} Sensitivity analysis of $\tau_r$ on the Enwik8 dataset. \textbf{a)} For the reference architecture with $N{=}1024$ and $d_m{=}15$ training remains stable and yields similar performance over an order of magnitude in $\tau_r$. Training runs with too large $\tau_r$ become unstable, while ones with too small timescale remove all signal from neuron output. \textbf{b)} The trade-off between number and complexity of neurons for the reference parameter budget of $P=3.29\cdot10^6$ yields the characteristic interior optimum irrespective of the tested timescales. However, large neuron configurations benefit from fast readout timescales, whereas small neuron configurations from slower ones, thereby shifting the exact location of the optimum.}
   \label{fig:enwik8_readout_tau_sensitivity_analysis}
\end{figure*}

\textbf{Choosing a good $\tau_r$ is simple,} as networks remain stable for a wide range of timescales (see Fig.~\ref{fig:enwik8_readout_tau_sensitivity_analysis}). We recommend a default of $\tau_r=5$ and sweeping the range [2,20]. On tasks like SHD-Adding, which have particularly sparse inputs yet long-range dependencies, choosing a larger $\tau_r$ can noticeably improve performance. Note that ELM neurons can still take advantage of orders of magnitude larger memory timescales $\tau_m$ than their high-pass filter timescale $\tau_r$, as in the case of SHD-Adding with 500 vs 5.

\subsection{Datasets and corresponding training details}
\label{sec:dataset_and_training_details}

\paragraph{Dataset and code availability.} SHD~\citep{cramer2020heidelberg} is available from Zenke Lab at \url{https://zenkelab.org/resources/spiking-heidelberg-datasets-shd/} under the Creative Commons Attribution 4.0 International License (CC BY 4.0). NeuronIO~\citep{beniaguev2021single} is available from David Beniaguev on Kaggle at \url{https://www.kaggle.com/datasets/selfishgene/single-neurons-as-deep-nets-nmda-train-data} under CC BY-SA 4.0. For SHD-Adding \citep{spieler2024the} and NeuronIO \citep{beniaguev2021single}, we use the dataloaders provided by Spieler et al. at \url{https://github.com/AaronSpieler/elmneuron}, released under the MIT License; the SHD-Adding dataloader ingests SHD data. Enwik8~\citep{mahoney2011large} is available from Matt Mahoney at \url{http://mattmahoney.net/dc/enwik8.zip}; it consists of the first $10^8$ bytes of the March 3, 2006 English Wikipedia dump and is licensed under the GNU Free Documentation License (GFDL). We preprocess and load Enwik8 using the standard Transformer-XL pipeline~\citep{dai2019transformer} provided by Dai et al. at \url{https://github.com/kimiyoung/transformer-xl}, released under the Apache License 2.0. 

\paragraph{Evaluations on the SHD-Adding dataset.} The SHD-Adding benchmark is based on the Spiking Heidelberg Digits dataset~\citep{cramer2020heidelberg}, and was introduced in~\citep{spieler2024the}. Each sample pairs two spike-encoded spoken digits (0--9, German or English) end-to-end, and the model must predict their sum (19 classes, chance $\approx 5.3\%$). The 700-channel binary spike trains are trimmed to one second per digit and discretized into 2\,ms bins, producing $2 \times 500 = 1000$ time steps. The network is optimized with cross-entropy on its final output. SuperSpike surrogate gradient used a scale of $100$. A 20\% validation split guides model selection, and test accuracy (mean $\pm$ std) is reported over three seeds. Where reported, reducible error denotes the mean test error minus a floor of $1 - 0.94^2 \approx 0.116$, based on a reference single-digit SHD accuracy of $0.94$~\citep{bittar2022surrogate}.

\begin{table}[h]
\centering
\caption{Training configurations for all three benchmarks.}
\label{tab:training_config}
\begin{tabular}{lccc}
\toprule
Setting & SHD-Adding & Enwik8 & NeuronIO \\
\midrule
\multicolumn{4}{l}{\textit{Optimization}} \\
\midrule
\quad Optimizer & Adamax & Adam & Adam \\
\quad Learning rate & 5e-3 & 5e-4 & 5e-4 \\
\quad LR schedule & cosine & cosine & cosine \\
\quad Warmup steps & 400 & 100 & 0 \\
\quad Gradient clip norm & 1.0 & 1.0 & 1.0 \\
\quad Batch size & 8 & 8 & 8 \\
\quad Epochs / turns & 70 & 750 & 35 \\
\quad Steps per epoch / turn & 2000 & 400 & 10000 \\
\quad Label smoothing & 1e-2 & 1e-4 & --- \\
\midrule
\multicolumn{4}{l}{\textit{Regularization}} \\
\midrule
\quad Input dropout & 0.2 & 0.0 & --- \\
\quad Recurrent dropout & 0.2 & 0.0 & --- \\
\quad MLP output L2 & 0.002 & 0.01 & --- \\
\quad Neuron activity L1 & 0.2 & 1.0 & --- \\
\bottomrule
\end{tabular}
\end{table}

\paragraph{Evaluations on the Enwik8 dataset.} Enwik8~\citep{mahoney2011large} is preprocessed following the standard Transformer-XL pipeline~\citep{dai2019transformer}. This yields 204 unique byte tokens, which we encode as one-hot vectors scaled by 3. The network processes sequences of length 100 and maintains context across consecutive batches through hidden state reuse: the recurrent hidden state is carried forward from one batch to the next, with a reset probability that decays from $1.0$ to $0.01$ over the first 40k training steps following a cosine schedule. At test time, evaluation proceeds sequentially with batch size 1 to ensure a single unbroken hidden state across the full test set. Models are trained using BPTT and cross-entropy loss; performance is reported as test bits-per-character (BPC $=$ loss $/\!\ln 2$), and the model with the lowest validation loss is selected for final evaluation. Where reported, reducible BPC denotes the remaining test BPC above the reference floor of $0.97$~\citep{rae2020compressive}.

\paragraph{Evaluations on the NeuronIO dataset.} The NeuronIO dataset~\citep{beniaguev2021single} provides simulated input-output data of a detailed biophysical layer 5 cortical pyramidal neuron model~\citep{hay2011models}, with 1278 binary pre-synaptic spike channels as input and somatic membrane voltage and output spikes as targets. Following~\citep{beniaguev2021single}, somatic voltage is capped at $-55$\,mV, offset by $-67.7$\,mV, and scaled by $1/10$. Each training sample spans 500\,ms, with the first 150\,ms excluded from the loss as burn-in. The model is a single ELM neuron trained using BPTT with equally weighted binary cross-entropy (spikes) and mean squared error (voltage). The model with the lowest validation RMSE is selected, and test voltage prediction performance is reported as reducible MAE (mean $\pm$ std) over three seeds. Where reported, reducible MAE denotes the test MAE minus a floor of $0.319$\,mV, the average achieved by the largest model in our sweep ($d_m=100$, AUC $= 0.9938$).

\paragraph{Fixed parameter budget experiments.} In case exactly matching the budget was not possible, the neuron count was rounded down for the $N$ vs $k_e$ sweeps, and the number of synapses per branch were rounded down for the $k_e$ vs $k_c$ experiments.

\paragraph{Activation regularization.} Two regularization terms are applied to the recurrent hidden layer. A per-neuron L2 penalty on the time-averaged absolute MLP output keeps individual neurons' memory updates away from $\tanh$ saturation, promoting gradient flow; this is particularly helpful for neurons with large $d_m$. A population-level L1 penalty on the mean neuron activity encourages sparse firing, particularly beneficial for wide layers. An ablation of these activity regularizers can be found in Appendix Fig.~\ref{fig:enwik8_regularizers_and_activity}. Strengths are listed in Table~\ref{tab:training_config}. Neither term is used for NeuronIO.

\subsection{Structured joint hyperparameter search experiment details}
\label{sec:structured_joint_hp_search_method_details}

\begin{table}[h]
\centering
\caption{Search space of the structured joint hyperparameter experiment.}
\label{tab:pareto_config}
\begin{tabular}{lccc}
    \toprule
    Setting & & SHD-Adding & Enwik8 \\
    \midrule
    \multicolumn{4}{l}{\textit{Swept settings}} \\
    \midrule
    \quad $N_\text{rec}$ & & \{32, 64, 128\} & \{512, 1024, 1536\} \\
    \quad $d_m$ & & \{3, 5, 10, 15\} & \{5, 10, 15, 20\} \\
    \quad $d_\text{mlp}$ & & \{$2d_m$, $3d_m$\} & \{$2d_m$, $3d_m$, $\lfloor 10\sqrt{d_m}\rfloor$\} \\
    \quad $d_\text{tree}$ & & \{$d_\text{mlp}$, $2d_\text{mlp}$\} & \{$d_\text{mlp}$, $2d_\text{mlp}$\} \\
    \quad $d_\text{branch}$ & & \{$d_\text{tree}/2$, $d_\text{tree}$\} & \{$d_\text{tree}$, $2d_\text{tree}$\} \\
    \midrule
    \multicolumn{4}{l}{\textit{Pareto candidate}} \\
    \midrule
    \quad $N_\text{rec}$ & & \{32, 64, 128, 256\} & \{256, 512, 768, 1024, 1280\} \\
    \quad $d_m$ & & $\lfloor\frac{1}{2}\sqrt{d_\text{inp}/15 + N_\mathrm{rec}}\rfloor$ & $\lceil\frac{1}{2}\sqrt{d_\text{inp} + N_\mathrm{rec}}\rceil$ \\
    \quad $d_\text{mlp},\; d_\text{tree},\; d_\text{branch}$ & & $2d_m,\; 2d_\text{mlp},\; d_\text{tree}$ & $2d_m,\; 2d_\text{mlp},\; d_\text{tree}$ \\
    \bottomrule
\end{tabular}
\label{tab:pareto_search_table}
\end{table}

\textbf{Additional configuration details.} The recurrence fraction is set to $\rho_\text{rec} = \sqrt{N_\text{rec}/(N_\text{rec}+d_\text{inp})}$ on SHD-Adding and $\rho_\text{rec} = N_\text{rec}/(N_\text{rec}+d_\text{inp})$ on Enwik8. On Enwik8, the readout layer neurons match the hidden layer in complexity and connectivity. On SHD-Adding, the readout layer connectivity is instead adjusted proportionally to $N_\text{rec}$ with $d_\text{tree} = d_\text{branch} \in \{10,15,20\}$. Regularization is switched to per-neuron scaling, with strengths (MLP~L2 at $10^{-5}$, activity~L1 at $10^{-3}$) selected to match the reference configuration's per neuron regularization strength at $N_\text{rec}=1024$; on Enwik8 the MLP~L2 term is applied to both ELM layers. The Pareto Configurations show mean and standard deviation across five seeds, otherwise, SHD-Adding results show mean across three seeds, whereas Enwik8 shows individual runs. Training otherwise follows the reference setup (Table~\ref{tab:training_config}).

\subsection{Theoretical framework experiments and data analysis}
\label{sec:theoretical_framework_data_analysis}

\begin{figure*}[ht]
   \centering
   \includegraphics[width=0.65\linewidth]{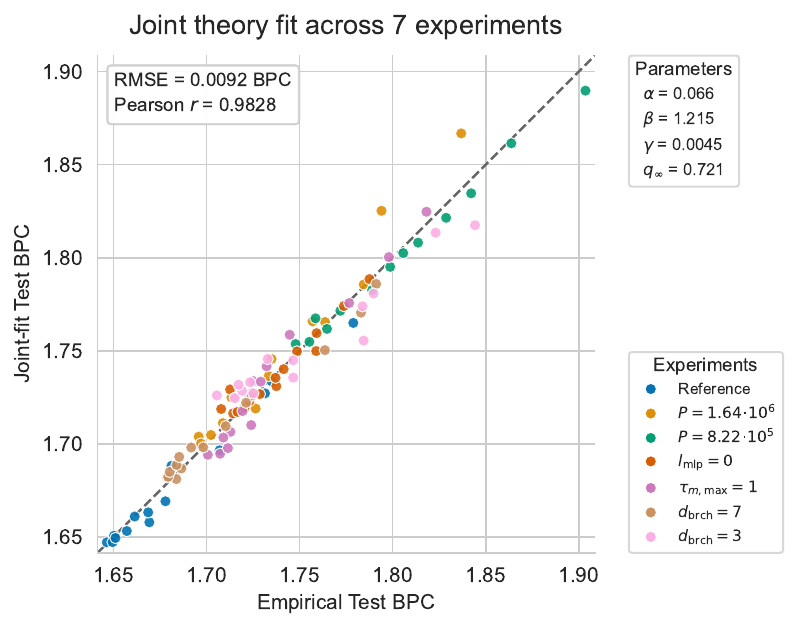} 
   \caption{\textbf{Joint theory-experiment fit across seven experiments.} Each point is one $(N_{\mathrm{rec}}, d_m)$ configuration; the seven experiments span three parameter budgets and multiple architectural correspondences to the two key theory parameters $\alpha$ ($\tau_{m,\mathrm{max}}$, $l_{\mathrm{mlp}}$) and $\beta$ ($d_{\mathrm{branch}}$). Eight theory parameters, the four shared reference values plus per-experiment versions for each varied quantity, and a single affine map are fit jointly to all 103 points by minimizing the sum of squared BPC using derivative-free optimization (differential evolution with L-BFGS-B polishing); reported $\alpha$, $\beta$, $\gamma$, $q_\infty$ refer to the reference architecture. A single parameter set captures both absolute BPC levels and the effects of architectural changes, supporting the view that theory parameters correspond to distinct identifiable architectural knobs.
   \textbf{Ablating} the core theory parameters $\alpha$ and $\beta$ that control how individual and collective neuron performance scales and trade-off against each other degrades the quantitative fit substantially ($r=0.7317$ and $r=0.4542$). Removing $q_\mathrm{inf}$ or setting $\gamma=1$ affects few points quantitatively ($r=0.9797$ and $r=0.9659$), but qualitatively degrade the fits by removing the convergence of curves for large $k_e$ and crossing of curves for different $\alpha$ for small $k_e$ configurations.\looseness-1
   }
   \label{fig:joint_theory_fit}
\end{figure*}

\begin{figure}[ht]
   \centering
   \includegraphics[width=0.8\linewidth]{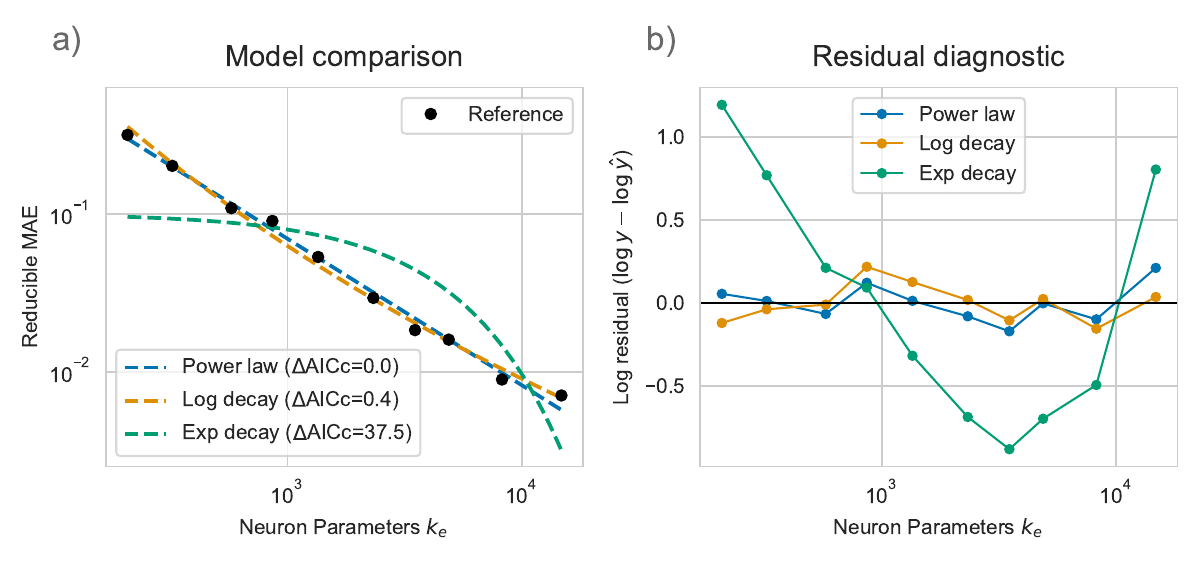}
    \caption{\textbf{A single neuron's reduction of approximation error is well described as a power law in parameter count.}
    Various sized ELM Neuron fit to NeuronIO with reducible MAE reported as mean over three seeds. Decay curves fitted in log-log.
    \textbf{a)} The reduction in error is compatible with a power law decay across two orders of magnitude in per-neuron parameters $k_e$, strongly preferred over exponential-decay alternatives ($\Delta\mathrm{AICc} = 37.5$), however, logarithmic decay cannot be ruled out ($\Delta\mathrm{AICc} = 0.4$).
    \textbf{b)} The unstructured residuals of the power law fit (compared to structured "U" shape of exponential fit) confirm the corresponding fit is approximately unbiased across the sweep.
    This finding motivates the per-neuron residual noise law $\sigma_n^2(k_e) \propto k_e^{-\alpha}$ used in the theory (Appendix~\ref{sec:eri-derivation}).\looseness-1}
   \label{fig:neuronio_powerlaw}
\end{figure}

\begin{figure*}[ht]
   \centering
   \includegraphics[width=1.0\linewidth]{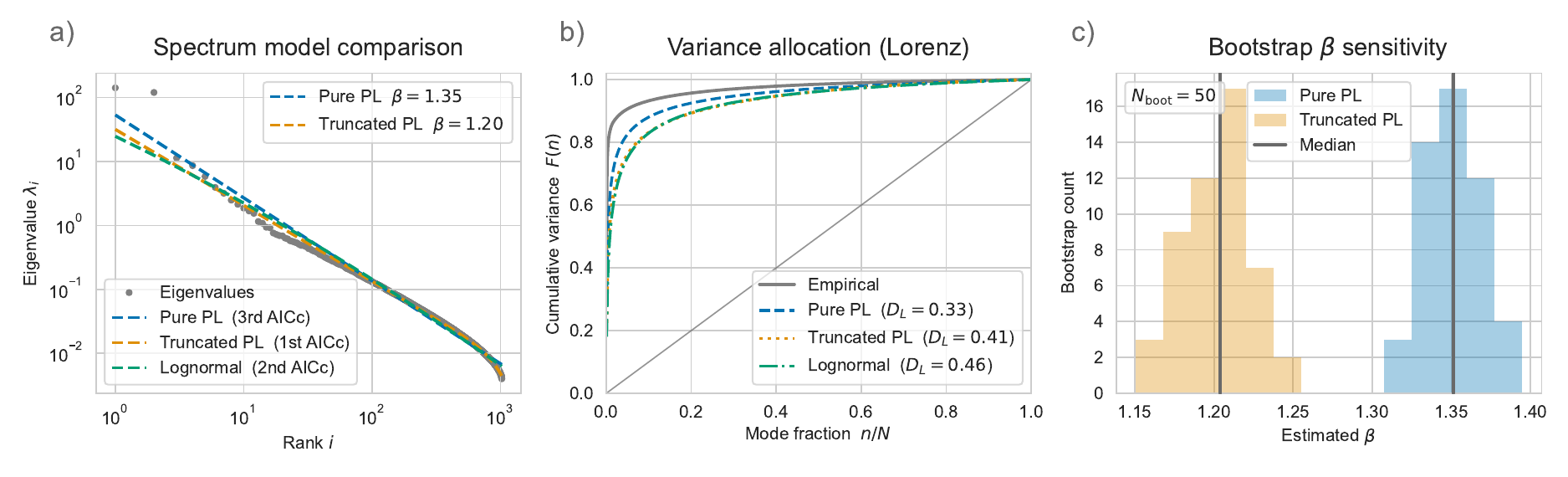}
    \caption{\textbf{Power-law structure in the ELM neurons' output.} Function fitting performed in log-log. Individual neuron output measured at memory readout $w^Tm_t$ as it contains the task-relevant slow signal components. Eigenvalues computed across 50 distinct recordings of 512 steps, after discarding 128-step burn-in, that were generated using the reference network with $N_\mathrm{rec}=1024$ evaluated on Enwik8.
    \textbf{a)} The neuron's output covariance eigenvalue spectrum is best captured by a truncated power law $\lambda_i = \sigma_f^2 \, i^{-\beta} \exp[-(i/i_c)^\nu]$ among the tested spectral models, supporting the assumed heavy-tailed spectral form.
    \textbf{b)} The Lorenz comparison provides an integrated check that the fitted spectrum preserves the variance allocation across modes, rather than only matching eigenvalues point-wise. 
    \textbf{c)} Bootstrap resampling of recordings with replacement ($N_\text{boot}=50$) for $\beta$ estimation quantifies the sensitivity of the fitted spectral exponent, with vertical lines marking the median $\beta$ values reported in panel a).
    \textbf{Note} that $\beta$ in our theory refers to the latent signal's power-law exponent; however, we can only directly measure the signal along with residual noise, which may bias the measurement. Notably, none of the measured spectra (even after output nonlinearity) show the flat floor that an isotropic additive residual would produce, indicating a structured rather than isotropic residual. However, since only the effective per-mode signal-to-noise ratios enter $I_\mathrm{rep}$, its form does not depend on how the residual is distributed across neurons. For more accurate estimates of $\beta$, the measurements may therefore use networks of highly expressive neurons that are assumed to have smaller approximation error (see also Fig.~\ref{fig:beta_invariance}).
    }
    \label{fig:enwik8_eigenvalue_spectrum}
\end{figure*}

Three analyses connect the theoretical proxy (Eq.~\ref{eq:I_rep}) to experimental observations. All curve fitting is performed by minimizing squared residuals in log-space via nonlinear least squares, with model selection by corrected Akaike Information Criterion (AICc). \\
\textbf{Measuring $\alpha$.} The per-neuron expressivity exponent $\alpha$ is estimated by fitting a power law to reducible error as a function of per-neuron parameters $k_e = \#\mathbf{w}_p + \#\mathbf{w}_r$. On NeuronIO ($d_m \in [2, 50]$), the metric is MAE $\propto \sigma_n \propto k_e^{-\alpha/2}$, where the halved exponent arises because MAE scales with the noise standard deviation rather than its variance (Fig.~\ref{fig:theory_introductory_figure}c, \ref{fig:neuronio_powerlaw}). On Enwik8 ($d_m \in [5, 35]$, $N_\text{rec}=1024$), $\alpha$ is recovered directly as the log-log slope of reducible BPC versus $k_e$, since in the low-SNR regime ($s \approx 1$, $d_m \leq 35$) the spectral exponent $\beta$ affects only the level of $I_\text{rep}$, not its scaling with $k_e$ (Appendix~\ref{sec:irep-scaling}; Fig.~\ref{fig:theory_scaling}g, Fig.~\ref{fig:enwik8_power_law}). Therefore, $\alpha$ slopes are best measured in the low-SNR regime, and $k_e$ sweeps should be truncated before the log bends or noise floor is reached. Measuring $\alpha$ beyond the low-SNR regime should be possible but would require a separate treatment and, in general, require estimating $\beta$ first. \\
\textbf{Measuring $\beta$.} The spectral decay exponent $\beta$ is estimated from the eigenvalue spectrum of the sample covariance of the memory readout $\mathbf{w}_r^\top \mathbf{m}_t$, centered across 50 batch trajectories of 512 time steps after discarding a 128-step burn-in, by fitting a truncated power law $\lambda_i = \sigma_f^2\, i^{-\beta}\exp[-(i/i_c)^\nu]$ with shared cutoff $(i_c, \nu)$ across different connectivity configurations (Fig.~\ref{fig:theory_scaling}h, \ref{fig:enwik8_eigenvalue_spectrum}). The memory readout was chosen as it represents the neuron output, which also contains the task-relevant slow components, more in line with the whole test set BPC evaluations. Measuring after the high-pass filter removes those slower components and artificially biases measurements towards significantly smaller $\beta$. \\
\textbf{Joint theory fit.} The joint theory fit maps $-I_\text{rep}$ to test BPC via a shared affine transformation across all seven Enwik8 parameter sweeps, using derivative-free global optimization (differential evolution) to minimize the sum of squared BPC residuals over eight free theory parameters (Fig.~\ref{fig:theory_scaling}a--f), with details in Fig.~\ref{fig:joint_theory_fit}. Some large $k_e$ configurations for performance degrading parameter setting $\tau_\mathrm{max}=1.0$ and $d_s$ / $50=7$ displayed unusually large deviations in performance compared to their neighboring configurations, and were rerun.

\newpage
\clearpage
\section{Derivation of the Effective Representation Information}
\label{sec:eri-derivation}

We derive the formula for the effective representation information,
\begin{equation}
  I_{\mathrm{rep}}(k_e) \;=\; \frac{1}{2}\sum_{i=1}^{P/(k_e + k_c)} \log_2\!\left(1 + s(k_e) \cdot i^{-\beta}\right),
\end{equation}
from the mutual information of a Gaussian vector channel, stating all assumptions. Under assumptions~\ref{A1}--\ref{A4}, the derivation in this section yields the expression for \(I_{\mathrm{rep}}(k_e)\) exactly.

\subsection{Assumptions}
\label{sec:assumptions}

\begin{enumerate}[label=\textbf{(A\arabic*)},leftmargin=3em]

\item \label{A1} \textbf{Gaussian channel.}
The layer output $\mathbf{y} \in \mathbb{R}^N$ decomposes as
$\mathbf{y} = \mathbf{f}(\mathbf{x}) + \mathbf{n}$, where
$\mathbf{x}\in\mathbb{R}^D$ is the input,
$\mathbf{f}:\mathbb{R}^D\to\mathbb{R}^N$ maps inputs to the task-relevant signal component,
and $\mathbf{n} \sim \mathcal{N}(\mathbf{0},\, \sigma_n^2 \mathbf{I}_N)$
is independent additive Gaussian noise. The signal is modeled as Gaussian:
$\mathbf{f}(\mathbf{x}) \sim \mathcal{N}(\mathbf{0},\, \mathbf{C}_f)$,
with positive-definite signal covariance $\mathbf{C}_f \in \mathbb{R}^{N \times N}$
matching the empirical second moments of the learned representation.

\item \label{A2} \textbf{Power-law signal spectrum.}
The signal covariance $\mathbf{C}_f$ has $N$ ordered eigenvalues $\lambda_1 \geq \lambda_2 \geq \cdots \geq \lambda_N > 0$ that follow a power law in the rank index: $\lambda_i = \sigma_f^2\, i^{-\beta}$, where the leading signal variance $\sigma_f^2 \equiv \lambda_1$ sets the scale and $\beta > 0$ is the spectral decay exponent.

\item \label{A3} \textbf{Phenomenological residual noise model.}
The total steady-state noise variance per neuron at per-neuron effective parameter count $k_e$ is modelled as
\[
  \sigma_n^2(k_e) \;=\; \sigma_f^2\, q(k_e), \qquad q(k_e) \;=\; \max\!\left((\gamma k_e)^{-\alpha},\; q_\infty\right),
\]
where $\gamma > 0$ is an effectivity constant, $\alpha > 0$ is the expressivity exponent, and $q_\infty > 0$ is an irreducible normalised residual floor set by the neuron's binding bottleneck --- most prominently its scalar output mechanism. Equivalently, the noise-to-signal variance ratio is assumed to decay as a power law in $k_e$ until pinned at this floor. This is adopted as a phenomenological ansatz for the effective residual mismatch of a finite-budget neuron.

\item \label{A4} \textbf{Fixed parameter budget.}
The total budget $P$ is divided equally among $N$ neurons, each receiving $P/N = k_e + k_c$ parameters: $k_e>0$ effective parameters driving expressivity and a connectivity overhead $k_c \geq 0$ that does not directly reduce approximation error.

\end{enumerate}

\subsection{Terminology}
\label{sec:terminology}

We evaluate the mutual information between the signal component and the layer's noisy output under the empirical data distribution. We formalise this quantity as the \emph{effective representation information}, $I_{\mathrm{rep}}(k_e)$. Unlike Shannon channel capacity---which requires maximising over all possible input distributions---$I_{\mathrm{rep}}(k_e)$ measures the information throughput achieved by the assumed representation channel under the network's architectural and finite-budget constraints.

\subsection{Derivation}
\label{sec:derivation-steps}

\paragraph{Step 1 (Mutual information of a Gaussian channel).}
Under~\ref{A1}, $I(\mathbf{f};\mathbf{y}) = h(\mathbf{y}) - h(\mathbf{y}|\mathbf{f}) = h(\mathbf{y}) - h(\mathbf{n})$. For Gaussian vectors, differential entropy is $\frac{1}{2}\log_2[(2\pi e)^N \det(\boldsymbol{\Sigma})]$; the prefactors cancel in the difference, and the ratio of determinants gives:
\begin{equation}\label{eq:mi-gaussian}
  I(\mathbf{f};\, \mathbf{y}) \;=\; \frac{1}{2}\log_2 \det\!\left(\mathbf{I}_N + \sigma_n^{-2}\,\mathbf{C}_f\right).
\end{equation}
\emph{Exact given~\ref{A1}.}\quad$\square$

Since the Gaussian distribution maximizes differential entropy for a fixed covariance, $h(\mathbf{y}) \geq h(\mathbf{y}_{\mathrm{true}})$ at matched second moments while $h(\mathbf{n})$ remains unchanged. Equation~\eqref{eq:mi-gaussian} therefore serves as an \emph{upper bound} on the mutual information of any \emph{additive channel with the same signal-independent noise distribution and the same output covariance}.

\paragraph{Step 2 (MIMO mutual information under power-law spectrum).}
Under~\ref{A2}, the matrix $\sigma_n^{-2}\,\mathbf{C}_f$ has eigenvalues $s\,i^{-\beta}$ for $i = 1, \ldots, N$, where $s = \sigma_f^2 / \sigma_n^2$ is the leading-mode signal-to-noise ratio. Using $\det(\mathbf{I} + \mathbf{A}) = \prod_i (1 + \lambda_i(\mathbf{A}))$, equation~\eqref{eq:mi-gaussian} becomes
\begin{equation}\label{eq:mi-sum}
  I(\mathbf{f};\, \mathbf{y}) \;=\; \frac{1}{2}\sum_{i=1}^{N} \log_2\!\left(1 + s\,i^{-\beta}\right).
\end{equation}
Each eigenmode contributes independently: mode $i$ carries $\frac{1}{2}\log_2(1 + s\,i^{-\beta})$ bits, with an effective per-mode signal-to-noise ratio of $s\,i^{-\beta}$ decreasing as a power law in the mode index. 

This spectral decay provides a natural soft cutoff on the number of informative dimensions. The effective dimensionality of the representation is governed by the stable rank of the signal covariance, $N_{\mathrm{eff}} = \sum_{i=1}^N i^{-\beta}$. Depending on the exponent $\beta$, $N_{\mathrm{eff}}$ either saturates ($\beta > 1$), grows logarithmically ($\beta = 1$), or grows sublinearly ($\beta < 1$). Consequently, high-order modes ($i \gg s^{1/\beta}$) contribute negligibly to the mutual information.

\emph{Exact given~\ref{A1},~\ref{A2}.}\quad$\square$

\paragraph{Step 3 (Per-neuron SNR from the phenomenological residual law).}
Under~\ref{A3}, the total noise variance at effective budget $k_e$ is
\[
  \sigma_n^2(k_e) \;=\; \sigma_f^2 \cdot \max\!\left((\gamma k_e)^{-\alpha},\; q_\infty\right).
\]
The leading-mode signal-to-noise ratio (SNR) is therefore
\begin{equation}\label{eq:snr}
  s(k_e) \;=\; \frac{\sigma_f^2}{\sigma_n^2(k_e)} \;=\; \min\!\left((\gamma k_e)^\alpha,\; q_\infty^{-1}\right).
\end{equation}
This is a power law in $k_e$ capped at a hard ceiling. It satisfies $s(k_e) = (\gamma k_e)^\alpha$ for small $k_e$ (parametric regime) and $s(k_e) = q_\infty^{-1}$ above the crossover threshold $k_e^* = \gamma^{-1}\, q_\infty^{-1/\alpha}$ (floor regime), where additional effective parameters provide no further increase in per-neuron SNR. Mode $i$ of the signal covariance thus experiences an effective SNR of $s(k_e)\,i^{-\beta}$.

\emph{Exact given~\ref{A3}.}\quad$\square$

\paragraph{Step 4 (Assembly).}
Substituting the layer width $N = P/(k_e + k_c)$ from~\ref{A4} and the SNR mapping $s = s(k_e)$ from~\eqref{eq:snr} into equation~\eqref{eq:mi-sum} yields:
\begin{equation}\label{eq:final}
  \boxed{\;I_{\mathrm{rep}}(k_e) \;=\; \frac{1}{2}\sum_{i=1}^{P/(k_e+k_c)} \log_2\!\left(1 \;+\; \min\!\left((\gamma k_e)^\alpha,\; q_\infty^{-1}\right) \cdot i^{-\beta}\right)\;}
\end{equation}
which completes the derivation.

\emph{Exact given~\ref{A1}--\ref{A4}.}\quad$\square$

\newpage
\clearpage

\subsection{Crossing of curves for varied \texorpdfstring{$\alpha$}{alpha} experiments}
\label{sec:gamma_crossing}

For a fixed total parameter budget $P$, varying the per-neuron budget $k_e$ also changes the layer width according to $N=P/(k_e+k_c)$. The dependence of the $I_{\mathrm{rep}}$ curve on the expressivity exponent $\alpha$ enters through the $s(k_e)=(\gamma k_e)^\alpha$ before the neuron noise floor. Since $1^\alpha=1$ for all $\alpha$, curves obtained by varying $\alpha$ will intersect where $\gamma k_e = 1$ (see Fig. \ref{fig:theory_scaling}b), specifically at
\[
k_\mathrm{cross}=\gamma^{-1},
\qquad
N_\mathrm{cross}=\frac{P}{\gamma^{-1}+k_c}.
\]
which permits direct estimation of $\gamma$ by identification of the crossing point $k_\mathrm{cross}$. A reduced $\gamma$ would shift the crossing to larger per-neuron budgets $k_\mathrm{cross}$ and, at fixed $P$, to smaller layer widths $N_\mathrm{cross}$.


\subsection{Scaling of \texorpdfstring{$I_{\mathrm{rep}}$}{Irep} with per-neuron budget}
\label{sec:irep-scaling}

We characterize how $I_{\mathrm{rep}}$ scales with per-neuron complexity~$k_e$ for a fixed layer size $N$, specifically when $s(k_e)$ is of order unity and pre noise floor saturation (the regime realized by most of our experiments). All $k_e$-dependence in equation~\eqref{eq:final} enters through the per-neuron SNR $s(k_e) = (\gamma k_e)^\alpha$, as $N = P/(k_e + k_c)$ is effectively $k_e$ independent due to the likewise growing budget $P$.

When $s$ is of order unity, the per-mode SNRs $s \cdot i^{-\beta}$ are small for all but the leading modes, and $\log_2(1+x) \approx x / \ln 2$ applies to the bulk of the sum, giving:
\begin{equation}\label{eq:irep-onset}
  I_{\mathrm{rep}}
  \;\approx\;
  \frac{(\gamma k_e)^\alpha}{2\ln 2}\sum_{i=1}^{N} i^{-\beta}
  \;\propto\; k_e^\alpha
\end{equation}
The log-log slope is therefore $\eta \approx \alpha$. The spectral exponent~$\beta$ enters only through the mode sum prefactor, setting the level of~$I_{\mathrm{rep}}$ but not its scaling with~$k_e$. This can be used to measure $\alpha$ directly from network layer performance, as was done in Fig.~\ref{fig:theory_scaling}g.


\subsection{Single-neuron noise-floor saturation}
\label{subsec:qinf_saturation}

The $I_{\mathrm{rep}}(k_e)$ expression (Eq.~\eqref{eq:final} implies two regimes for the neuron-level SNR: an increasing regime for smaller per-neuron budgets $k_e$ and a noise-floor-limited regime once $(\gamma k_e)^\alpha$ reaches the value $1/q_\infty$, as seen from the expression for $s(k_e)$:
\[
s(k_e)=\min\!\left((\gamma k_e)^\alpha,\frac{1}{q_\infty}\right).
\]

\begin{figure*}[ht]
   \centering
   \includegraphics[width=0.75\linewidth]{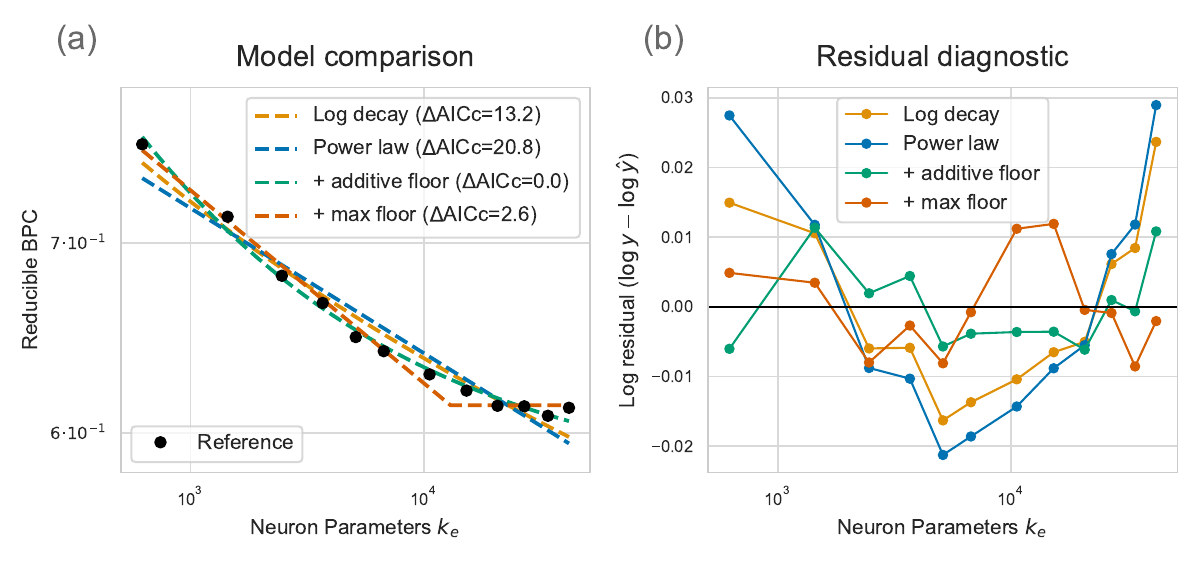}
   \caption{\textbf{Empirical neuron complexity scaling on Enwik8 independently validates the max noise floor assumption:}
    Reducible test BPC vs. per-neuron effective parameter count $k_e$, with layer width kept fixed at $N=1024$ but increasing layer budgets. Functions fitted in log-log, with floor function slopes seeded with a pure power-law fit, and floors seeded with the last evaluation point. We compare models using the corrected Akaike Information Criterion (AICc).
    \textbf{a)} The additive floor and max floor $\max(c k_e^{-\alpha}, f)$ models are strongly preferred over the pure power law or log decay (see $\Delta$AICc in the legend),  consistent with Assumption~\ref{A3}, but are not significantly distinguishable from each other.
    \textbf{b)} While residuals reveal a bump for the max floor fit just around its knee (marking the transition into the noise floor), compared to the log decay and pure power law fits (and somewhat additive floor) it doesn't seem to diverge towards the noise floor.}
    \label{fig:enwik8_power_law}
\end{figure*}

The transition between these regimes occurs where the two branches are equal and is empirically observable as a knee in a fixed-layer-width neuron-complexity scaling curve (see Fig.~\ref{fig:enwik8_power_law}).
Solving for $q_\infty$ at the transition point $k_{\mathrm{sat}}$ gives
\[
q_\infty=(\gamma k_{\mathrm{sat}})^{-\alpha}.
\]
which permits direct estimation of the single-neuron noise-floor parameter using independently determined values of $\alpha$ and $\gamma$.

\newpage
\clearpage
\section{Additional Results and Supporting Evidence}
\label{sec:additional_results_supporting_evidence}

\begin{figure*}[ht]
   \centering
   \includegraphics[width=0.65\linewidth]{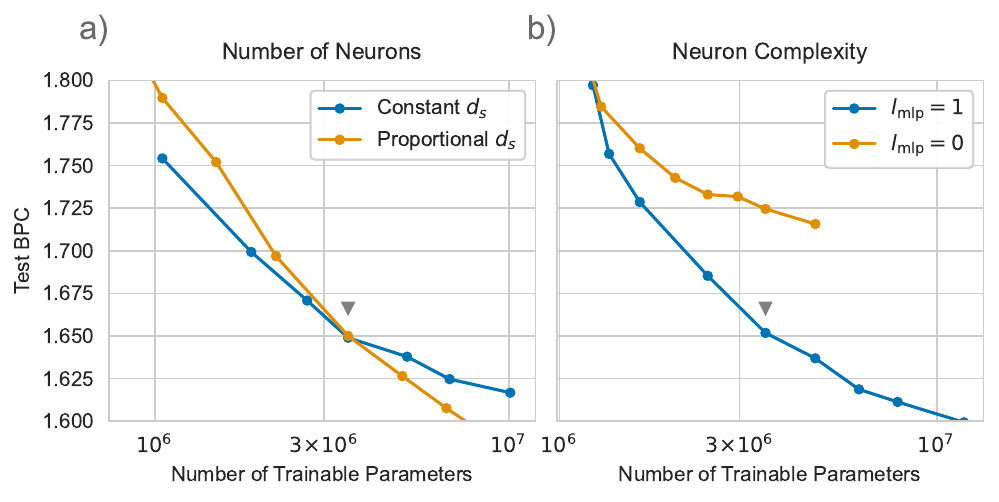}
   \caption{\textbf{Proportional connectivity and deeper dendritic integration scale better even accounting for additional parameter cost:}
   experimental sweep across number of neurons and neuron complexity on Enwik8 matching Figure \ref{fig:enwik8_empirical_results}a,b, with x-axis plotting in terms of total network trainable parameters.
   While curves rescale, the same trend emerges; 
   \textbf{a)} scaling with number of neurons works better with proportional neuron connectivity, and
   \textbf{b)} scaling with neuron complexity with hierarchical synaptic integration beats simpler integration by a large margin.
   }
   \label{fig:enwik8_neuron_network_scaling_params_matched}
\end{figure*}

\begin{figure}[t]
    \centering
    \includegraphics[width=0.9\textwidth]{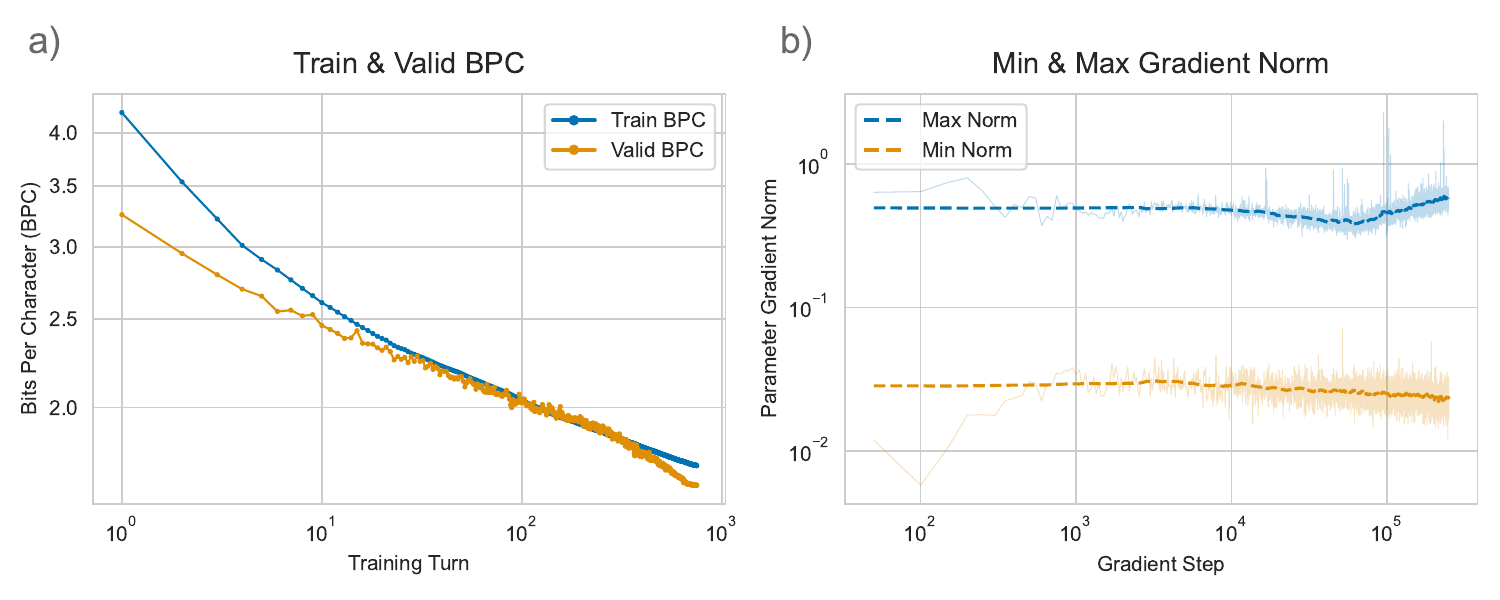}
    \caption{\textbf{ELM Network training is smooth and gradients remain stable throughout:} Training dynamics of the reference run with $N{=}1024$ and  $d_m{=}15$ in Figures~\ref{fig:theory_scaling}. \textbf{a)}~Train and valid BPC over 750 training turns, converging near $1.644$ valid BPC slightly below test BPC of $1.65$. \textbf{b)}~Min and max parameter gradient norms, logged every 50th gradient step for $5000$ samples, remain stable throughout training with no signs of exploding or vanishing gradients.}
    \label{fig:training_diagnostics}
\end{figure}

\begin{figure*}[ht]
   \centering
   \includegraphics[width=0.85\linewidth]{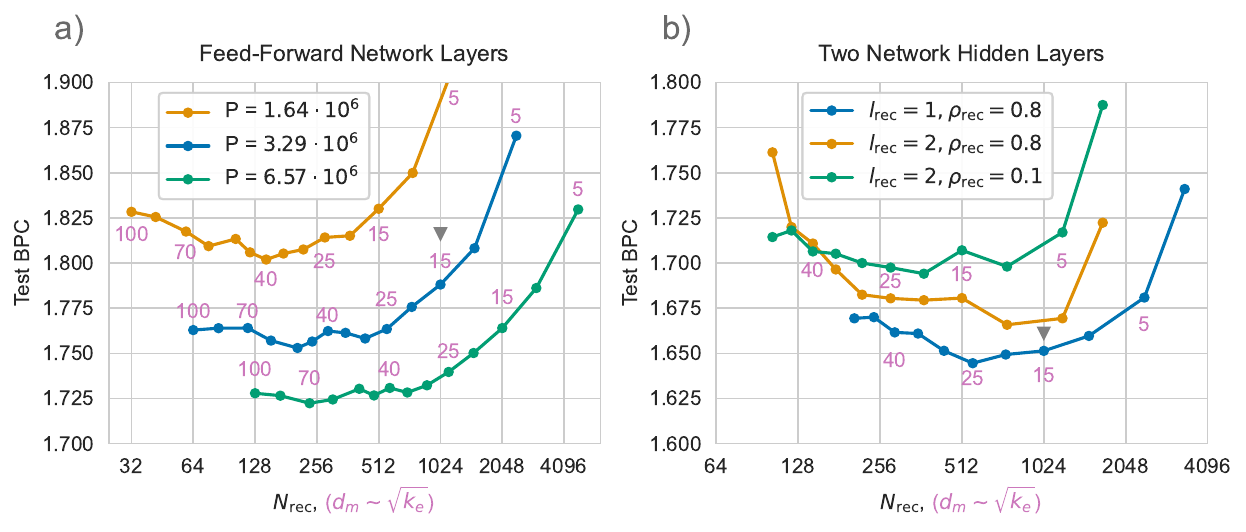}
   \caption{\textbf{The qualitative trade-off between number of neurons and complexity also persists for feed-forward and two-hidden-layered networks:}
   We evaluate the $N$ vs $k_e$ tradeoff on Enwik8 for purely feed-forward ELM-Network with $\rho_\mathrm{rec}=0.0$ and networks with an additional hidden layer but same parameter budget. Note that individual ELM Neurons remain internally recurrent irrespective of $\rho_\mathrm{rec}$, and for parameter matched networks with twice the hidden layers the per-hidden-layer budget is halved.
   \textbf{a)} We likewise observe the emergence of an interior optimum, that shift towards more and more complex neurons with increasing budget like in Figure~\ref{fig:enwik8_empirical_results}. Overall network performance is degrading substantially ($\sim$0.1 test BPC) and the optimal neuron complexity has increased significantly, yet the optimum is not as sharp anymore as before. The network possibly relies more on individual neurons extracting meaningful features on their own, unable to build on recurrent features.
   \textbf{b)} The neuron complexity vs neuron count tradeoff is preserved, with the optimum shifting towards smaller neuron configurations, as our theoretical proxy would predict for such network layers with half parameter budget. Performance also degrades overall (additional vertical shift), consistent with a reduced effectivity of the network parameters, and the optimum sharpened somewhat. Reducing recurrence results in the same qualitative change in performance as in single hidden layer networks.\looseness-1
   }
   \label{fig:enwik8_scaling_in_ff_and_2hidden_networks}
\end{figure*}

\begin{figure*}[ht]
   \centering
   \includegraphics[width=0.8\linewidth]{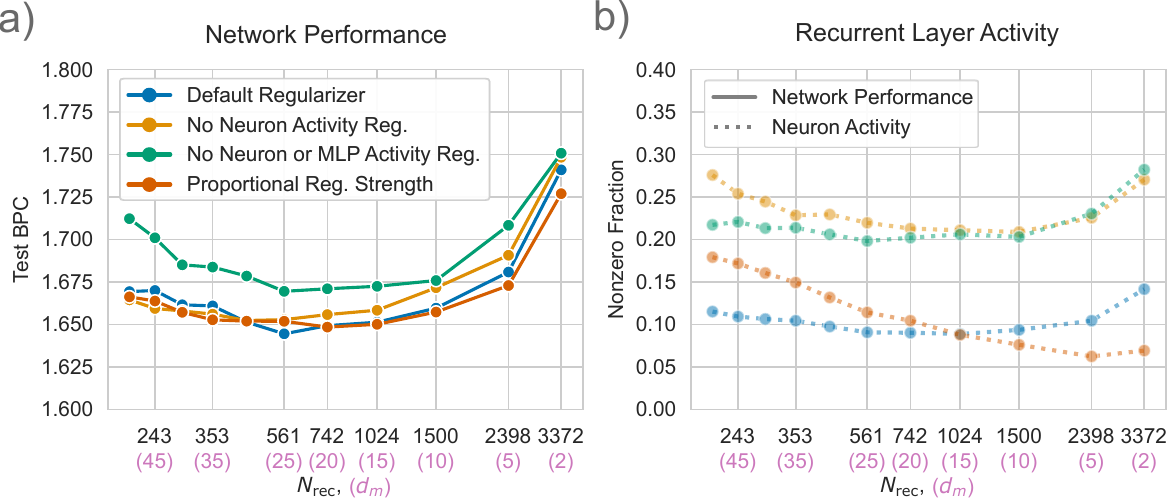}
   \caption{\textbf{The ELM-Network benefits from L2 neuron MLP output and L1 network activity regularization, without changing the qualitative tradeoffs:}
   Regularizer setup described in Appendix \ref{sec:dataset_and_training_details}.
   \textbf{a)} Compared to the default regularizer setup (blue curve), disabling the L1 regularizer on network activity (orange curve) slightly degrades performance for larger networks. Additionally disabling the L2 regularizer on the neurons MLP output more strongly degrades performance for particularly large neuron configurations. Scaling both regularizer strength automatically based on number of neurons, slightly improves performance at the parameter sweep extremes.
   \textbf{b)} The corresponding average recurrent layer activations measured across 2000 steps. Disabling the L1 neuron output regularizer roughly doubles network activity. Additionally disabling the L2 MLP output regularizer results in reduced activity levels for networks with large neuron configurations, as some neurons become unstable. Number of neuron proportional activity regularization improves performance across the board, now delivering consistent per neuron regularization strength across network configurations, further dropping network activity for wide networks while simultaneously improving performance. 
   }\label{fig:enwik8_regularizers_and_activity}
\end{figure*}

\begin{figure*}[ht]
   \centering
   \includegraphics[width=1.0\linewidth]{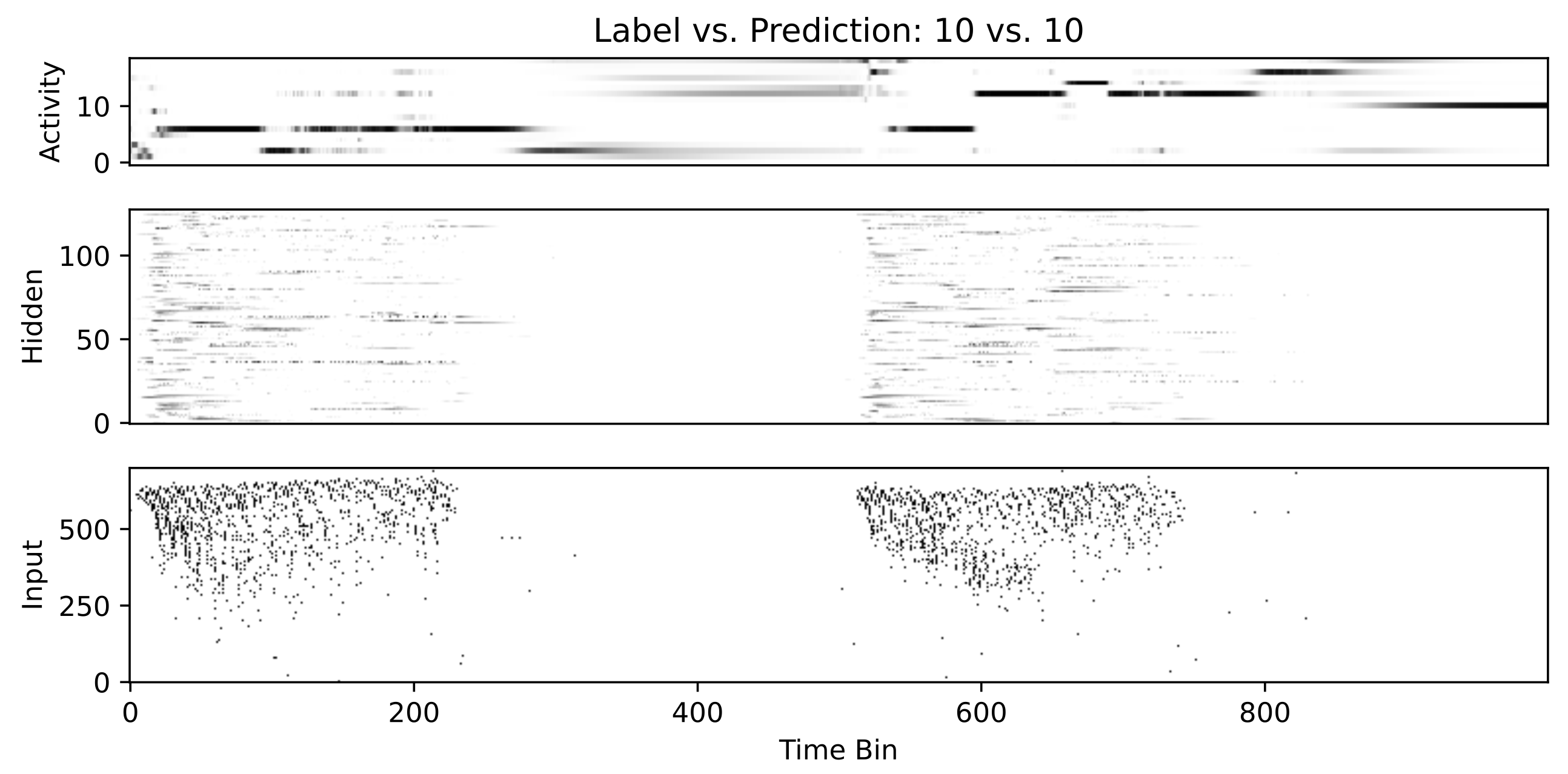}
   \caption{\textbf{The ELM-NET architecture exhibits rich temporal dynamics, characterized by periods of asynchronous irregular firing, synchronized bursts, and network silence:}
   An example network inference of an ELM Network with 128 neurons on SHD-Adding (see Appendix \ref{sec:dataset_and_training_details}).
   The networks' hidden layer displays short spike or burst like activity, with more active network phases visibly correlating to high input activity periods where individual digits were spoken. In between digit presentation the network falls mostly silent.
   The readout layer ELM neuron remains active throughout all periods, and displays frequent confident switching between different predictions; ultimately settling correctly.
   }\label{fig:shd_data_visualization}
\end{figure*}

\begin{figure*}[ht]
   \centering
   \includegraphics[width=1.0\linewidth]{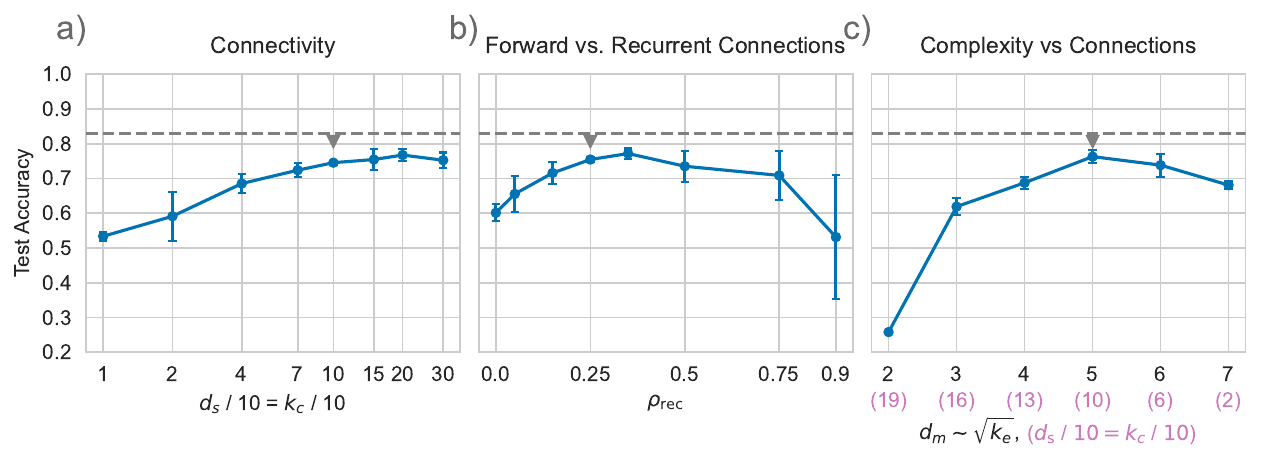}
   \caption{\textbf{Connectivity introduces the same qualitative tradeoffs on SHD-Adding as on Enwik8:}
    Estimated dataset noise floor at $83\%$ marked with dashed line.
    \textbf{a)} Increasing synapses per branch $d_{\mathrm{branch}}$ improves performance with diminishing returns.
    \textbf{b)} An optimal recurrent fraction $\rho_{\mathrm{rec}} \approx 0.30$ exists, which is roughly proportional to the ratio of recurrent to recurrent plus input connections.
    \textbf{c)} The tradeoff between neuron complexity and neuron connectivity shows an optimum for balanced configurations.
    These tradeoffs qualitatively mirror the Enwik8 findings (Fig.~\ref{fig:enwik8_empirical_results}d--f).}
    \label{fig:shd_connectivity_appendix}
\end{figure*}

\begin{figure*}[ht]
   \centering
   \includegraphics[width=1.0\linewidth]{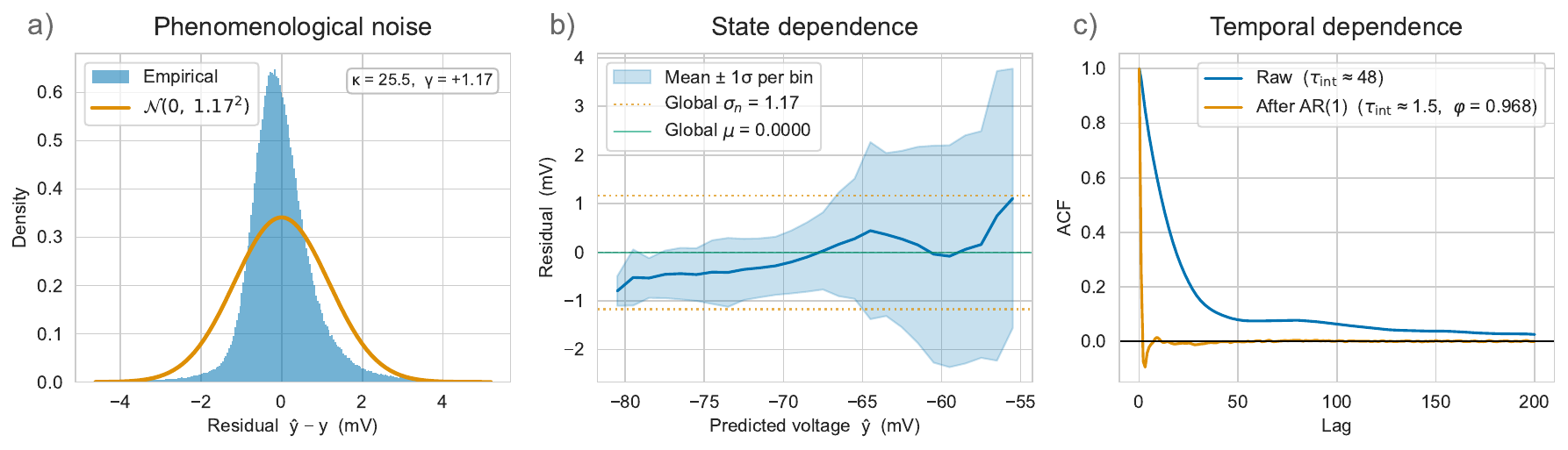}
   \caption{\textbf{The ELM neuron voltage prediction residuals on NeuronIO have some state and temporal dependence:}
   Residuals for a neuron model with $d_m=3$ fitted on the NeuronIO dataset containing single neuron voltage recordings. Note that the underlying target membrane voltage data itself displays multiple operating regimes, with particularly violent dynamics towards spiking threshold ($\approx-60mV$).
   \textbf{a)} Voltage-prediction residuals are concentrated near zero with mild skew and a sharper-than-Gaussian peak, suggesting the additive noise model to be reasonable, with deviations that can be treated as phenomenological noise.
   \textbf{b)} Residual mean and one-standard-deviation band per predicted-voltage bin: the standard deviation grows mildly toward depolarized states, indicating state-dependent heteroskedasticity. 
   \textbf{c)} Residual autocorrelation function: the raw residuals are temporally correlated ($\tau_{\mathrm{int}} \approx 48$), but an AR(1) correction with $\varphi = 0.968$ collapses the ACF to near-zero ($\tau_{\mathrm{int}} \approx 1.5$), showing the temporal structure is dominated by a single slow mode.}
   \label{fig:neuronio_gaussianity}
\end{figure*}

\begin{figure*}[ht]
   \centering
   \includegraphics[width=0.75\linewidth]{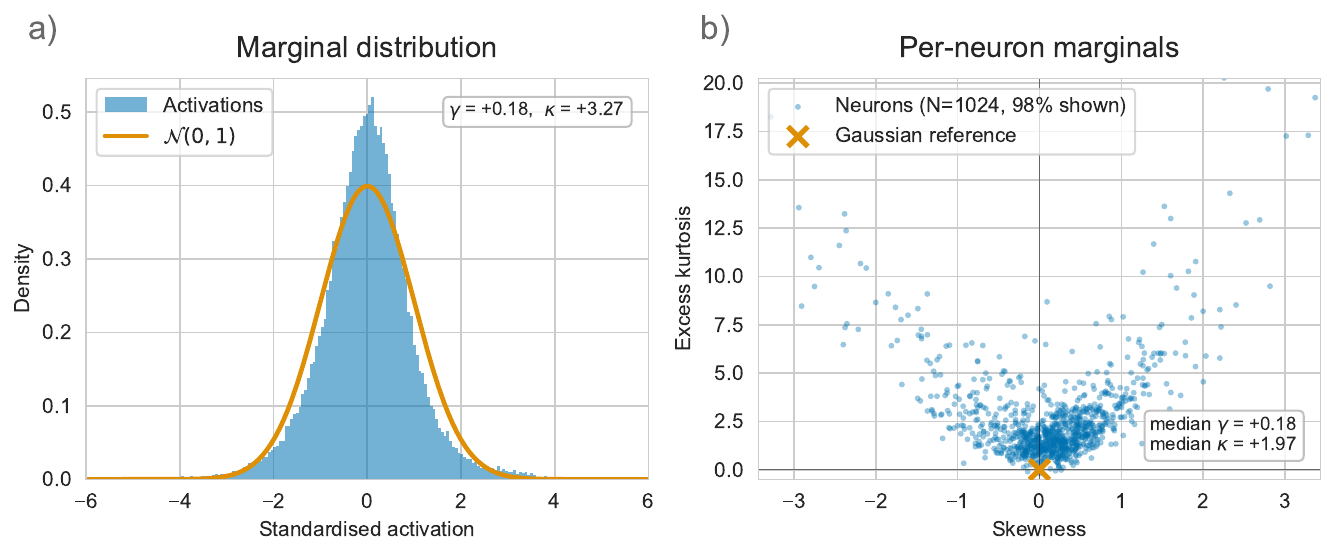}
   \caption{\textbf{Gaussianity of the readout $w_r^\top m_t$ of neurons on Enwik8.}
    Only $y = f + n$ is observable, so Gaussianity is tested on $y$ directly. Data plotted from 50 distinct recordings of 512 steps, after discarding 128-step burn-in. Recordings from the reference model with $N_\mathrm{rec}=1024$ on Enwik8.
    \textbf{a)} Pooled marginal of per-neuron $z$-scored activity vs. $\mathcal{N}(0,1)$, shape-only. \textbf{b)} Per-neuron (skew, excess
    kurtosis) with Gaussian reference at $(0,0)$.}
   \label{fig:enwik8_gaussianity}
\end{figure*}

\begin{figure*}[ht]
    \centering
    \includegraphics[width=1.0\linewidth]{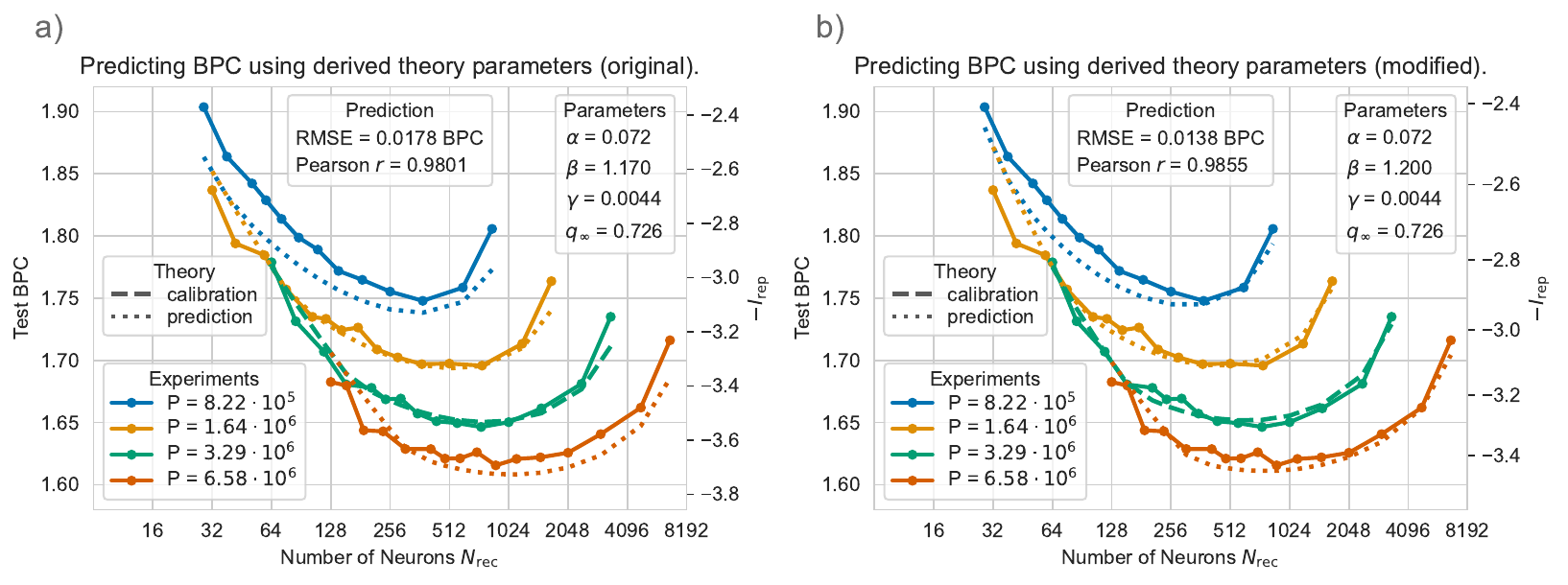}
    \caption{\textbf{Independently derived theory parameters predict how the optimal neuron-number--complexity tradeoff shifts across unseen parameter budgets.}
    This analysis tests whether the theoretical proxy can extrapolate across budgets when its parameters are obtained independently rather than jointly fitted to the displayed tradeoff curves.
    \textbf{Parameter derivation:}
    The neuron-expressivity exponent $\alpha=0.072$ is measured from the pre-saturation scaling of reducible BPC with per-neuron complexity $k_e$ at fixed layer width (Fig.~\ref{fig:theory_scaling}g and Appendix Fig.~\ref{fig:enwik8_power_law}a, explanation Appendix \ref{sec:irep-scaling}).
    The spectral exponent $\beta=1.17$ is measured from a truncated power-law fit to the recurrent-layer activity covariance spectrum (Fig.~\ref{fig:theory_scaling}h and Appendix Fig.~\ref{fig:enwik8_eigenvalue_spectrum}, details in Fig.~\ref{fig:enwik8_eigenvalue_spectrum}).
    The common crossing of the changed $\alpha$ curves is visually identified near the final point of the reference series in Fig.~\ref{fig:theory_scaling}b, at $k_e=225$, giving $\gamma=k_{\mathrm{cross}}^{-1}=1/225\approx0.0044$ (explanation in Appendix~\ref{sec:gamma_crossing}.
    The saturation knee is likewise identified visually in Appendix Fig.~\ref{fig:enwik8_power_law}a: the fourth point from the right is selected because performance no longer improves beyond it, yielding $q_\infty=0.726$ through $q_\infty=(\gamma k_{\mathrm{sat}})^{-\alpha}$ (explanation in Appendix~\ref{subsec:qinf_saturation}).
    \textbf{Mapping calibration:}
    The $I_{\mathrm{rep}}$-to-test-BPC mapping is calibrated using a single experiment series at the reference budget, $P=3.29\cdot10^6$, shown by dashed theory curves.
    After calibration, the mapping is held fixed and used to predict the tradeoff curves at all other budgets, shown by dotted curves.
    Solid curves show empirical test BPC, and the reported metrics are pooled across the predicted budgets.
    \textbf{a)}
    With the affine mapping $\mathrm{BPC}=b+aI_{\mathrm{rep}}$, the locations of the optima and their shifts with budget are predicted qualitatively and quantitatively ($r = 0.9801$) quite well.
    Nonetheless, BPC at the quarter budget is slightly underestimated for small $N$, while performance at the doubled budget is somewhat overestimated for large $N$.
    \textbf{b)} With a saturating exponential mapping $\mathrm{BPC}=B_{\min}+a\exp(-cI_{\mathrm{rep}})$ with floor $B_{\min}=0.97$ (which can account for diminishing performance gains caused by the fixed-size readout layer, as well as finite task-performance limits not explicitly represented by $I_{\mathrm{rep}}$) and a $\beta=1.20$ (a slightly higher measured median of the bootstrap estimates in Appendix Fig.~\ref{fig:enwik8_eigenvalue_spectrum}c) alignment for large-$N$ is improved and under and over predictions reduced for quarter and double budgets. 
    More robust $\beta$ estimation, e.g., by averaging measurements across multiple trained models, might improve performance further.}
   \label{fig:theory_predictions}
\end{figure*}

\clearpage
\null\vfill
\begin{figure*}[ht]
   \centering
   \includegraphics[width=0.85\linewidth]{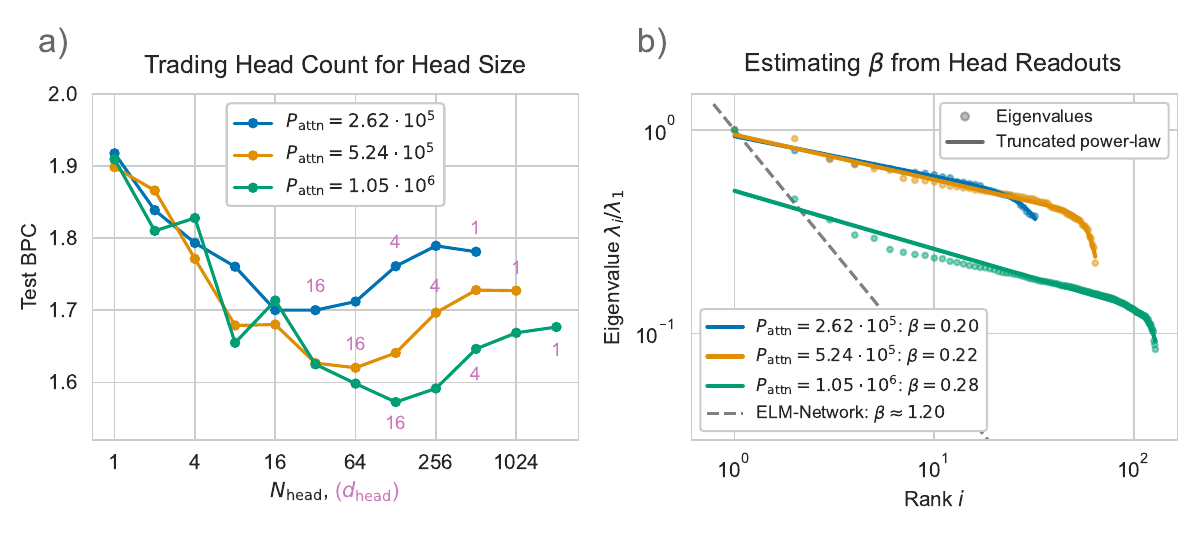}
   \caption{\textbf{Similar unit count vs complexity tradeoffs emerge in multi-head attention Transformers, with higher head-count being preferred as budget grows, consistent with a flatter activity spectrum.}
   Analogous to experiment Fig.~\ref{fig:enwik8_empirical_results}c we train a range of Transformers on Enwik8 and sweep the number of attention heads $N_\mathrm{head}$ against the per-head dimension $d_\mathrm{head}$ (see also \citep{vaswani2017attention}) at three fixed attention parameter budgets $P_\mathrm{attn}$ determined by $(N_\mathrm{head} \cdot d_\mathrm{head}) \in \{512,1024,2048\}$ being held fixed respectively.
   We use a pre-LN decoder-only Transformer \citep{radford2019language} of three blocks with $d_\mathrm{model} = 128$ and dense feed-forward width $4 d_\mathrm{model}$, otherwise trained as in Appendix~\ref{sec:dataset_and_training_details}. 
   Secondary $d_\mathrm{head}$ axis labels in pink.
   \textbf{a)} A clear interior optimum emerges at every budget and shifts towards more heads with it, as in Fig.~\ref{fig:enwik8_empirical_results}c. Unlike in ELM-Networks, however, the optimal unit-size remains roughly constant at $d_\mathrm{head} \approx 16$ across all three budgets.
   \textbf{b)} We consider each head's additive write into the residual stream as its output activity. As these are vector-valued, we reduce each pair's cross-covariance to its trace, which keeps one channel per head. Eigenvalue spectra of the covariance across head outputs, measured and fitted as in Appendix~\ref{sec:theoretical_framework_data_analysis}, yield $\beta \approx 0.20$--$0.28$, far below the $\beta \approx 1.20$ of the ELM Network's recurrent layer. Having a vector-valued output possibly helps different heads to occupy more distinct activity dimensions.
   \textbf{Interestingly}, with $N_{\mathrm{eff}} = \sum_i i^{-\beta}$ (Appendix~\ref{sec:derivation-steps}), the measured $\beta$ places the attention layer in the sublinearly growing regime. Additional heads therefore keep contributing informative dimensions, and the substantial inter-unit redundancy that drives the optimum towards more complex units as in Fig.~\ref{fig:enwik8_empirical_results}c does not set in across the swept budgets and number of heads, consistent with our theoretical proxy's prediction for $\beta \ll 1$. 
   \textbf{Furthermore}, a simple unit complexity sweep using $d_\mathrm{head}$ at fixed $N_\mathrm{head} = 32$ and otherwise unchanged hyperparameters yields an estimate of $\alpha \approx 0.122$, slightly higher than the ELM-Networks of $\alpha \approx 0.072$, also indicating better expressivity scaling.
   \textbf{Note} that $d_\mathrm{head}$ corresponds only roughly to the per-unit complexity $k_e$, so part of it arguably acts as connectivity overhead $k_c$ instead, in line with the caveat raised in Section~\ref{sec:discussion_limitations}. The $\beta$ estimates carry the same signal-plus-noise bias discussed in Fig.~\ref{fig:enwik8_eigenvalue_spectrum}.\looseness-1}
   \label{fig:transformer_head_tradeoff}
\end{figure*}
\vfill

\begin{figure*}[ht]
    \centering
    \includegraphics[width=1.0\linewidth]{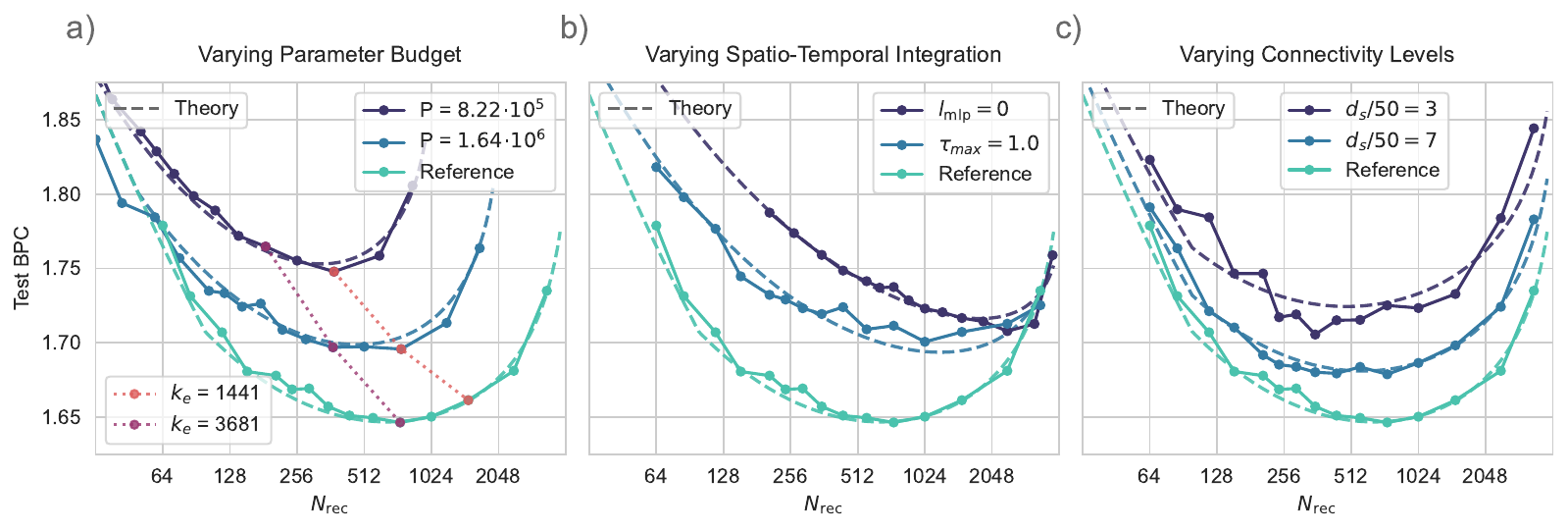}
    \caption{\textbf{The theoretical proxy model qualitatively reproduces the empirical scaling tradeoffs.}
    The displayed experimental tradeoff results and corresponding jointly fitted $I_\mathrm{rep}$ based theory curves (dashed lines) are the same as in Fig.~\ref{fig:theory_scaling}a-f) but plotted on top of each other. Note that the theory curves extend somewhat beyond the experimentally probed range.
    While the joint theoretical fit tracks the empirical results across all 7 experiments quite well overall, smaller discrepancies remain. For example, the $ q_\mathrm {inf} $- caused knee and its corresponding steeper performance degradation (left side of each subplot) might happen too late and is likely too steep, indicating that possibly an earlier but softer saturation $q_\mathrm{inf}$ mechanism would be more accurate.
    }
    \label{fig:theory_scaling_overlay}
\end{figure*}

\begin{figure*}[ht]
    \centering
    \includegraphics[width=1.0\linewidth]{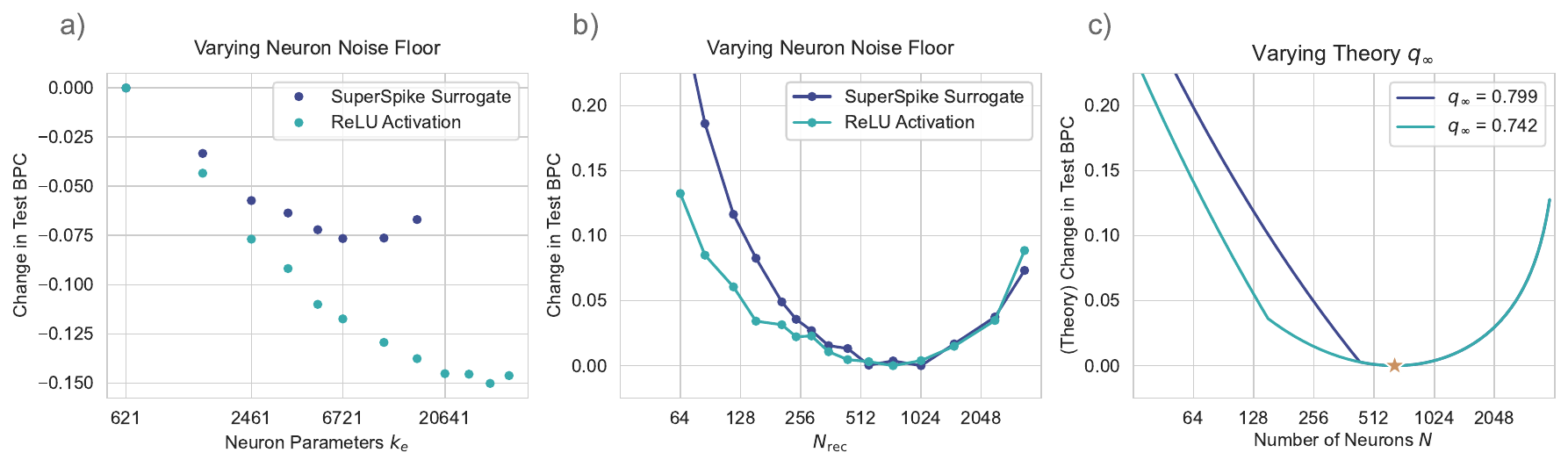}
    \caption{\textbf{Changing to binary neuron activation increases neuron noise floor $q_\infty$ and is qualitatively matched by the theoretical proxy.}
    All results for ELM Networks are evaluated on Enwik8. Change in Test BPC is plotted throughout to exclude a noise-floor-unrelated effect of SuperSpike that leads to an overall degradation in performance (vertical shift, possibly related to change in $\gamma$).
    \textbf{a)} Sensitivity analysis with respect to neuron parameters $k_e$ for ELM Networks with $N_\mathrm{rec}=1024$ and either default ReLU or surrogate SuperSpike neuron activation functions. The noise floor is directly visible as the point above which no meaningful improvement in performance happens, which is roughly at 20k and 6.7k parameters per neuron for ReLU and SuperSpike, respectively, indicating a higher neuron noise floor for SuperSpike networks. Particularly large neuron configurations in combination with less robust training methods can lead to training instabilities resulting in a decrease in performance (e.g., see last point SuperSpike).
    \textbf{b)} Empirical evaluations for the fixed budget neuron count vs neuron complexity experiment with both activation functions. Mostly, the performance for small numbers $N_{\mathrm{rec}}$ of complex neurons is affected (left side of the panel), conceptually consistent with large neurons reaching their noise floor, possibly with training instabilities causing a widening performance gaps for very large neuron configurations.
    \textbf{c)} Corresponding theory proxy curves with  $q_\infty$ values derived using the formula in Appendix \ref{subsec:qinf_saturation} and $k_\mathrm{sat}$ values $6721$ and $20641$ respectively from panel a). Overall, the qualitative behavior of the experiment is reproduced, although the theoretical proxy postulates a sharper transition than observed (see also the smoother noise-floor transition in panel a).\looseness-1
    }
    \label{fig:theory_qinf_comparison}
\end{figure*}

\begin{figure*}[ht]
    \centering
    \includegraphics[width=0.85\linewidth]{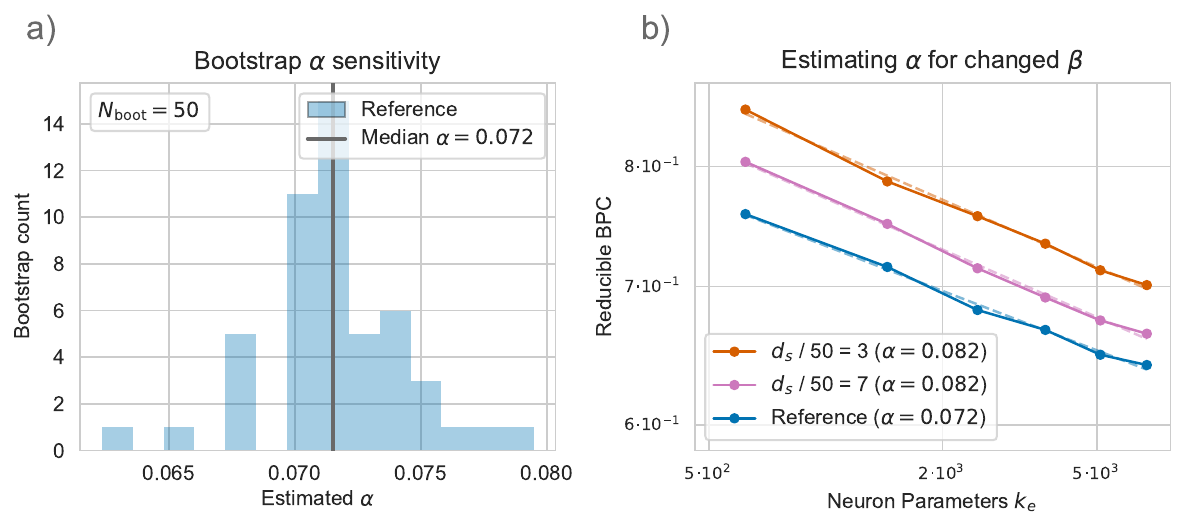}
    \caption{\textbf{The neuron-expressivity exponent $\alpha$ measurements shows smaller bootstrap spread, and smaller sensitivity to connectivity changes, compared to changes induced by spatio-temporal integration capabilities.}
    Measurements for ELM Networks with $N_\mathrm{rec}=1024$ and $d_m=15$ for panel a) and increasing $d_m$ for panel b).
    \textbf{a)} Bootstrap resampling of the observed $(k_e,\mathrm{BPC})$ pairs with replacement ($N_{\mathrm{boot}}=50$) also yields the single reference measurement of $\alpha=0.072$. The bootstrap spread is smaller relative to the separation between the reference and the direct spatio-temporal-integration modifying experiment measurements in Fig.~\ref{fig:theory_scaling}g which yielded $0.036$ and $0.044$ respectively.
    \textbf{b)} Reducing connectivity, which should conceptually only impact the independently measured $\beta$, yields matching $\alpha$ estimates within the approximate bootstrap spread. The deviations are also much smaller than changes induced by directly weakening neuron integration in Fig.~\ref{fig:theory_scaling}g, which yielded $\alpha$ estimates of $0.036$ and $0.044$ respectively. This further evidences a reasonably clean association between theory and architectural parameters.}
    \label{fig:alpha_experiments}
\end{figure*}

\begin{figure*}[ht]
    \centering
    \includegraphics[width=1.0\linewidth]{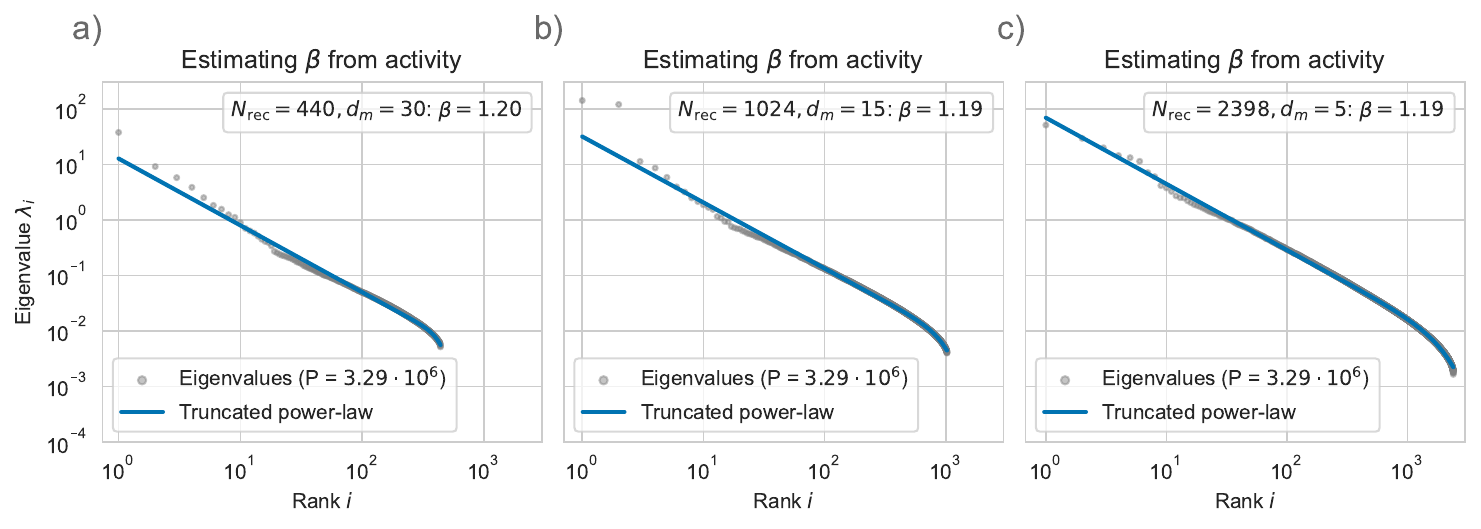}
    \caption{\textbf{The spectral exponent $\beta$ is invariant, within bootstrap spread, across network width and neuron complexity at a fixed parameter budget.}
    Same activity recording and power-law fitting protocol as in Fig.~\ref{fig:enwik8_eigenvalue_spectrum} (cutoff fitted per configuration), applied to three configurations of the reference budget sweep $P=3.29\cdot10^6$ of Fig.~\ref{fig:theory_scaling}a.
    \textbf{a--c)} The three network configurations yield roughly the same $\beta = 1.20$, $1.19$, $1.19$, well within the bootstrap spread of Fig.~\ref{fig:enwik8_eigenvalue_spectrum}c, despite an order-of-magnitude change in per-neuron parameters $k_e$ and a fivefold change in width, however, show slight deviation from the shared-cutoff measurement yielding $1.17$ in Fig.~\ref{fig:theory_scaling}h, likewise within the spread. 
    The framework treats $\beta$ as a layer property shared along a budget sweep irrespective of neuron count or neuron complexity; the invariance observed here is consistent with this, supports the shared-$\beta$ assumption behind the joint fit and the budget extrapolation of Fig.~\ref{fig:theory_predictions}, and indicates that $\beta$ can be measured reliably within a given neuron type.
    \textbf{Note}, however, that preliminary measurements of $\beta$ across widely changed $\alpha$ (such as in Fig.~\ref{fig:theory_scaling}g) were less consistent, likely due to the measured activity spectrum containing signal and residual together (see note in Fig.~\ref{fig:enwik8_eigenvalue_spectrum}), and architectural changes also altering the residual structure. We overall recommend measuring $\beta$ within networks of neurons with sufficient spatio-temporal integration capability (higher $\alpha$), such as the reference configuration.\looseness-1
    }
    \label{fig:beta_invariance}
\end{figure*}

\begin{figure*}[ht]
    \centering
    \includegraphics[width=0.55\linewidth]{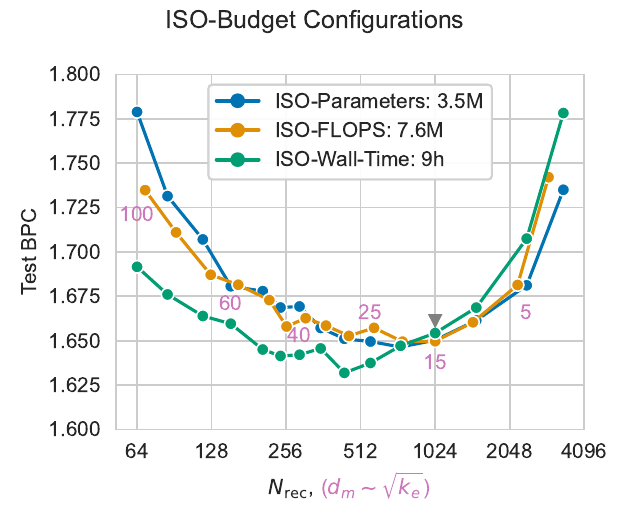}
    \caption{\textbf{The characteristic tradeoff between number of neuron and complexity of neuron persists across different budget types:} besides (trainable) parameters we additionally consider (inference) FLOPS and (training) wall-time of the whole network as budgets. 
    For ISO-FLOPS (7.6M for single step) large neuron configurations gain width whereas small neuron configurations lose width, yet the tradeoff essentially looks qualitatively and quantitatively unchanged
    For ISO-Wall-Time (9h) the interior-optimum shifts towards larger neuron network configurations, which execute faster (resulting in more training steps compared to simpler neuron configurations) due to increased hardware alignment of larger but fewer matrix operations on NVIDIA GPUs. The resulting varied training length might be accounted for using varied $\gamma$ in the theoretical proxy.
    }
    \label{fig:iso_budget_ablations}
\end{figure*}

\end{document}